\documentclass[conference]{IEEEtran}
\IEEEoverridecommandlockouts

\usepackage{cite}
\usepackage{amsmath,amssymb,amsfonts}
\usepackage{algorithmic}
\usepackage{graphicx}
\usepackage{enumerate}
\usepackage{subfigure}
\usepackage{textcomp}
\usepackage{amsmath}
\usepackage{url}
\usepackage{multirow}
\newtheorem{dfn}{Definition}
\def\BibTeX{{\rm B\kern-.05em{\sc i\kern-.025em b}\kern-.08em
    T\kern-.1667em\lower.7ex\hbox{E}\kern-.125emX}}
\begin{document}
%\history{Date of publication xxxx 00, 0000, date of current version xxxx 00, 0000.}
%\doi{10.1109/ACCESS.2024.0322000}

\title{Survey of Privacy Threats and Countermeasures in Federated Learning}

\author{\IEEEauthorblockN{Masahiro HAYASHITANI, Junki MORI, and Isamu TERANISHI}
\IEEEauthorblockA{\textit{Secure System Platform Research Laboratories, NEC Corporation}\\
hayashitani@nec.com, junki.mori@nec.com, teranisi@nec.com}
}

\maketitle

\begin{abstract} %サーベイが必要な理由を追記
%% V1
% Federated learning is widely considered a privacy-aware learning method because no training data is exchanged directly between clients. Nevertheless, there are threats to privacy in federated learning, and privacy countermeasures have been studied. However, we note that common and unique privacy threats have been categorized and described only in horizontal federated learning. In this paper, we describe privacy threats and countermeasures by categorizing threat models and privacy attacks among three typical types of federated learning (horizontal federated learning, vertical federated learning, and federated transfer learning).

%% V2
 Federated learning (FL) has emerged as a privacy-preserving machine learning paradigm that enables collaborative model training without directly exchanging raw data among clients. While FL mitigates the privacy risks associated with centralized data collection, it remains vulnerable to various privacy threats that can compromise sensitive information. Existing literature has only focused on privacy threats and countermeasures within horizontal federated learning (HFL) and vertical federated learning (VFL) individually. This paper provides a comprehensive and systematic review of privacy threats across all three principal FL paradigms: HFL, VFL, and federated transfer learning (FTL). We introduce a unified taxonomy that categorizes privacy threats according to the data type targeted, and we discuss corresponding defense mechanisms.
\end{abstract}


\maketitle

\section{Introduction} %サーベイが必要な理由を追記
% 本サーベイの意義
% 連合学習のプライバシーリスクを、HFL, VFL, FTLの三つの観点から体系的に整理した
% 特に、プライバシーリスクとなるデータの対象によって6つに攻撃を分類したのは本サーベイが初めて
% FTLに関しては、プライバシーリスクがまとめられていなかった
% HFLとVFLは個別にはリスクが議論されていた
% 本サーベイによって、どのタイプの連合学習を使用するかで考慮すべきプライバシーリスクと、それに対して取るべき対処法の指針となる

% イントロの流れ

% 第一パラグラフ
% 連合学習が重要になっている。
% データの形式に応じて3つのタイプの連合学習がある

% 第二パラグラフ
% すべてのタイプの連合学習は、直接データをやり取りしないため、プライバシーを考慮した手法ではあるが、まだリスクはある。
% HFLやVFLなど個々の連合学習についてはプライバシーリスクが調査されサーベイ論文としてもまとめられているが、FTLも含めたすべてのタイプの連合学習のリスクと対処法を体系的にまとめた論文はない。
% 本論文では、すべてのタイプの連合学習のプライバシーリスクを包括的に、リスクの対象となるデータに応じて6つの攻撃に分類した。
% 本サーベイによって、どのタイプの連合学習を使用するかで考慮すべきプライバシーリスクと、それに対して取るべき対処法の指針となる

\label{sec:introduction}
%% V4
With the increasing ubiquity of computing devices, individuals generate substantial volumes of data daily. The centralized collection and storage of such data are significantly resource-intensive and time-consuming \cite{lyu2020threats}. Furthermore, collecting user data raises significant privacy and confidentiality concerns, as the data often includes sensitive personal information. In light of growing societal awareness of privacy issues, regulatory frameworks such as the European Union’s AI Act have emerged, further limiting the viability of large-scale data aggregation practices.

In response to these challenges, federated learning (FL) has become a privacy-preserving machine learning paradigm that enables multiple clients to train a global model collaboratively without sharing their raw local data. FL methodologies are typically classified into three categories based on the alignment of data features and samples across clients: horizontal federated learning (HFL), vertical federated learning (VFL), and federated transfer learning (FTL).

These approaches are widely regarded as promising for privacy preservation, as they mitigate the need to expose individual-level raw data. Nevertheless, FL remains susceptible to privacy risks. Various studies have demonstrated that communicating model updates during training can inadvertently leak sensitive information to other clients, the central server, or external adversaries \cite{lyu2020threats}.

While the privacy risks inherent to specific forms of FL such as HFL and VFL have been examined and summarized in prior survey studies (e.g., \cite{lyu2020threats} for HFL and \cite{liu2024vertical} for VFL), there remains a lack of a unified and systematic review encompassing all paradigms of FL, including FTL. This paper presents a comprehensive overview of privacy risks across all FL variants, categorizing them into six distinct types of privacy threats based on the nature of the data targeted. We have searched papers on privacy threats and countermeasures in all paradigms of FL. The survey guides the privacy risks and corresponding mitigation strategies that need to be considered for each FL setting.

\section{Categorization of Federated Learning}
\begin{figure*}[tb]
    \centering
    \subfigure[Horizontal federated learning.]{\includegraphics[width=0.31\textwidth]{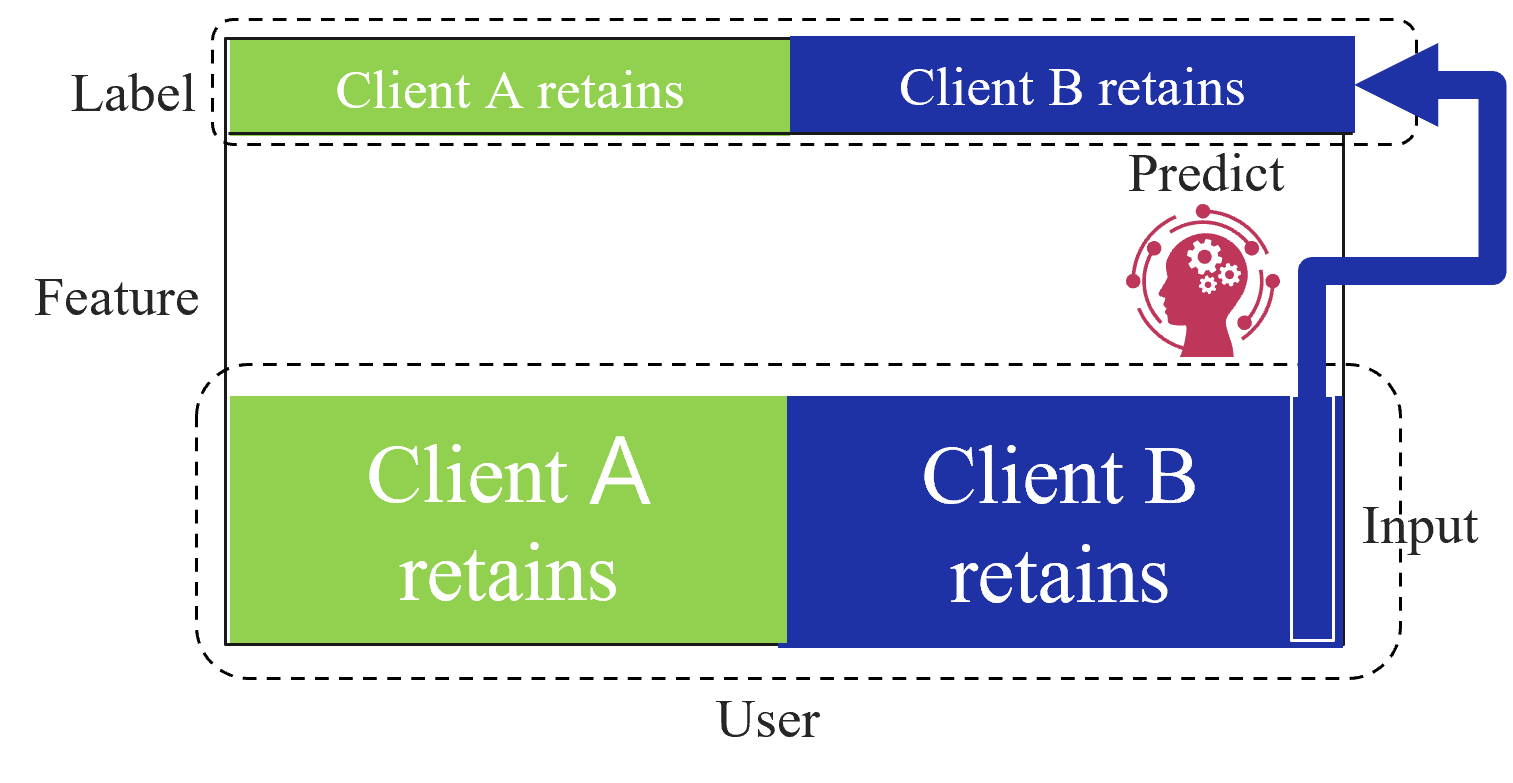}\label{HFL_structure}}
    \subfigure[Vertical federated learning.]{\includegraphics[width=0.31\textwidth]{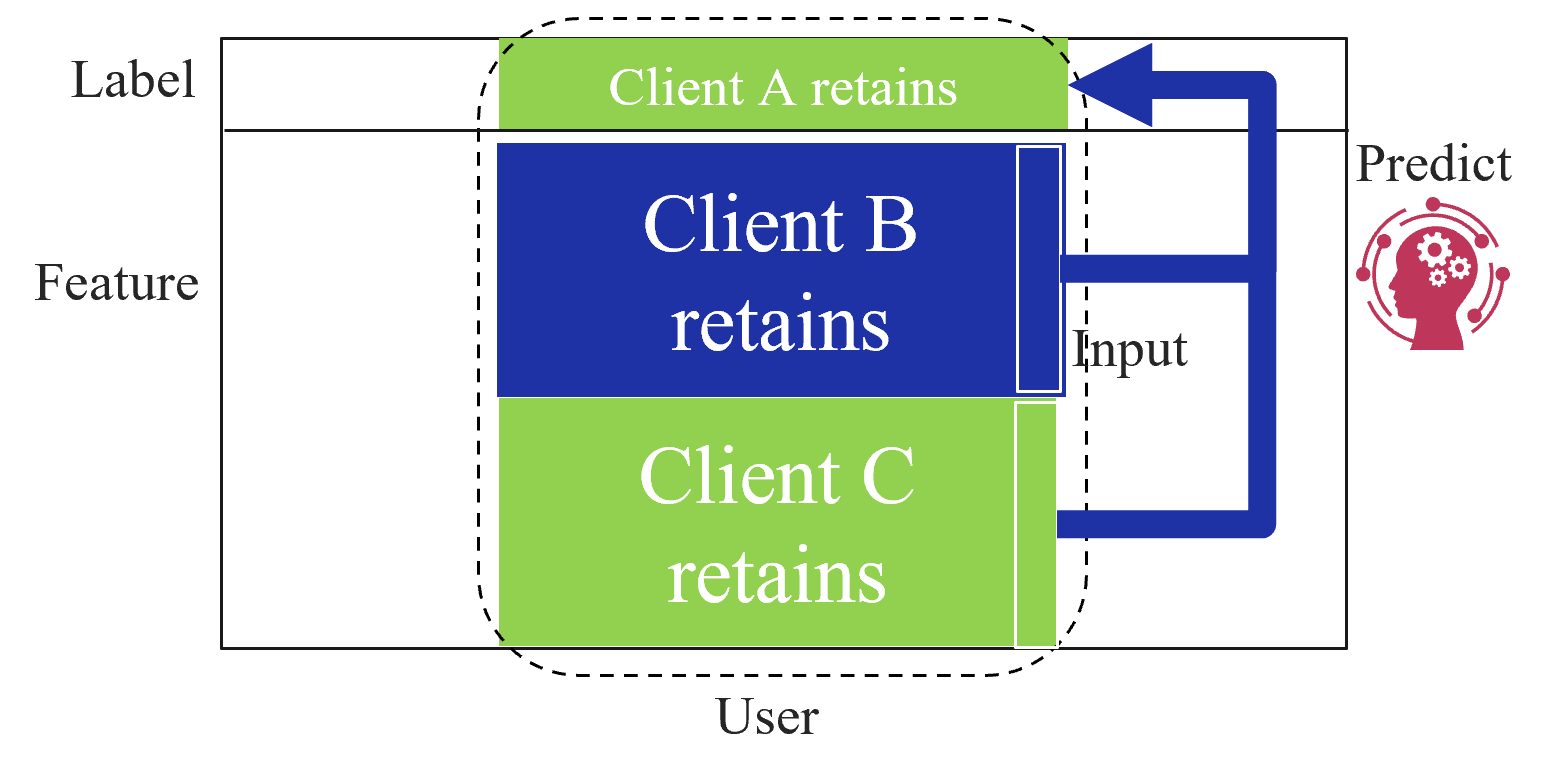}\label{VFL_structure}}
    \subfigure[Federated transfer learning.]{\includegraphics[width=0.31\textwidth]{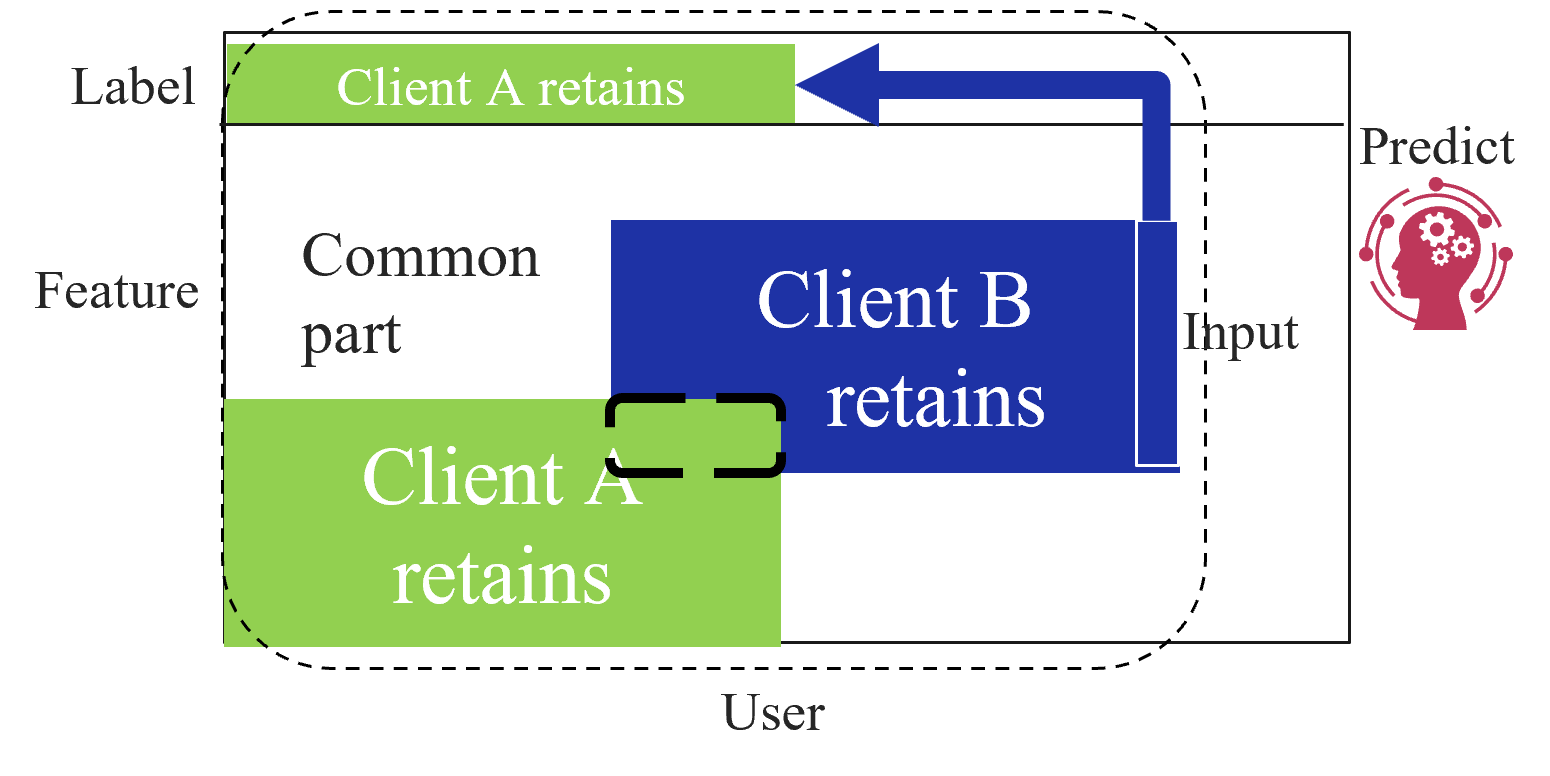}\label{FTL_structure}}
    \caption{Categorization of federated learning based on data structure owned by clients.} \label{FL_category}
\end{figure*}

We first review three types of FL based on the data structures among clients, as introduced by Yang et al. \cite{yang2019federated}: HFL, VFL, and FTL. Figure \ref{FL_category} shows each type's data structure among clients. HFL assumes that each client has the same features and labels but different samples (Figure \ref{HFL_structure}). In contrast, VFL assumes that clients share the same samples but possess disjoint features (Figure \ref{VFL_structure}). FTL addresses cases where clients differ in both samples and features (Figure \ref{FTL_structure}).

The following subsections describe the learning and prediction methods for each FL type.

\subsection{Horizontal Federated Learning}
HFL is the most common federated learning category that Google first introduced \cite{mcmahan2017communication}. The goal of HFL is for each client to hold different samples and collaboratively improve the accuracy of a model with a common structure. 

\begin{figure}[tb]
\begin{center}
\includegraphics*[width=70mm]{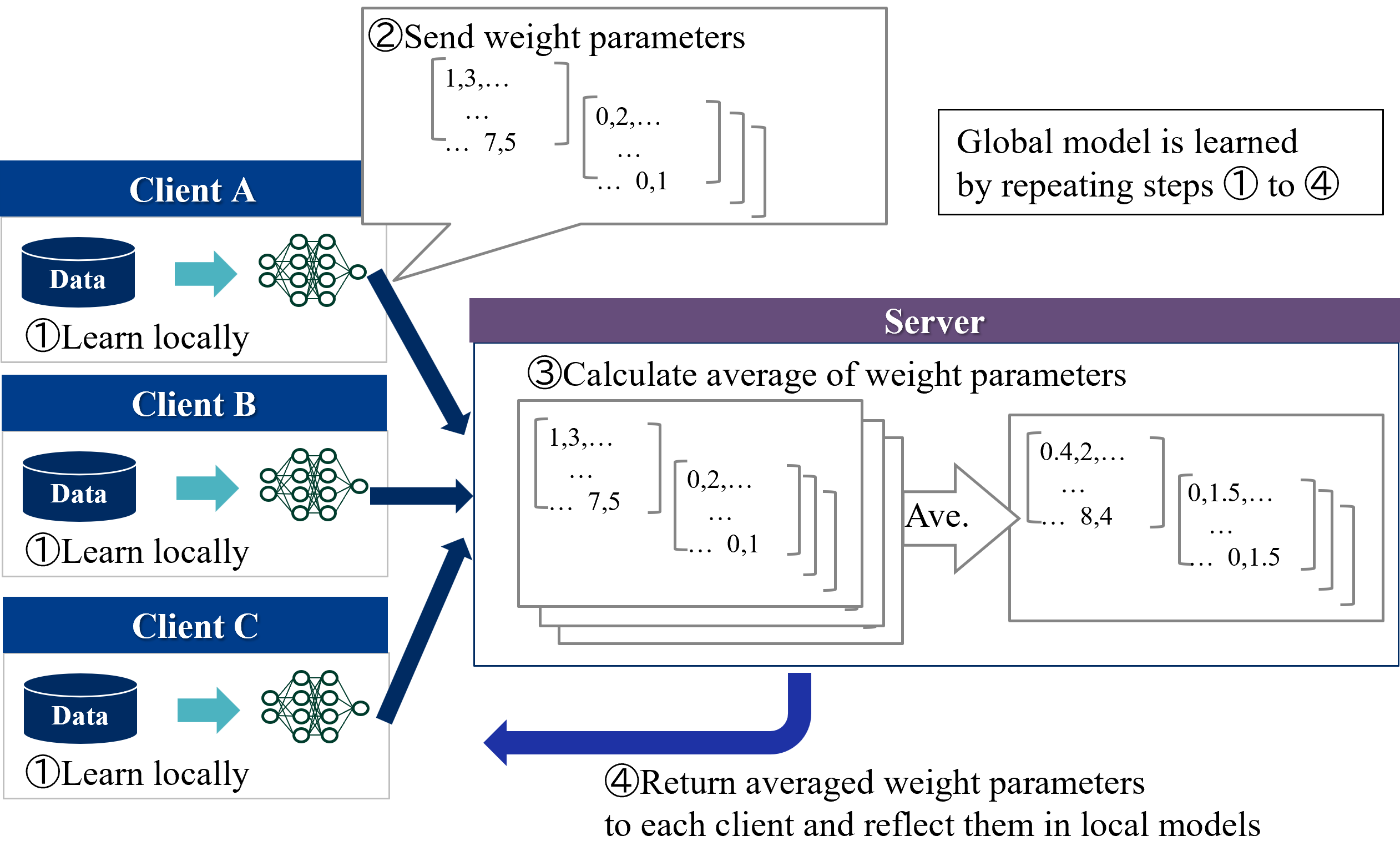}
\end{center}
\caption{Overview of the HFL learning protocol.}
\label{LM_HFL}
\end{figure}

Figure \ref{LM_HFL} shows an overview of the HFL learning protocol. Two types of entities participate in learning of HFL:

\begin{enumerate}
    \renewcommand{\labelenumi}{\roman{enumi}}
    \item \textbf{Server} - Coordinator. Server exchanges model parameters with the clients and aggregates model parameters received from the clients.
    \item \textbf{Clients} - Data owners. Each client locally trains a model using their own private data and exchanges model parameters with the server.
\end{enumerate}
Each client first trains a local model for a few steps and sends the model parameters to the server. The server then updates the global model by aggregating the local models, typically by averaging, as in FedAvg, and distributes the result to all clients. This process repeats until convergence. During inference, each client independently predicts labels using the global model and its features.

This protocol is called centralized HFL because it relies on a trusted third-party server. In contrast, decentralized HFL eliminates the central server, allowing clients to communicate directly, thereby reducing communication costs \cite{beltran2023decentralized}.

% Each clients first trains a local model for a few steps and sends the model parameters to the server. Next, the server updates a global model by aggregating (in standard methods such as FedAvg, simply averaging) the local models and sends it to all clients. This process is repeated until the convergence. During inference time, each client separately predicts the label using a global model and its own features.

% The protocol described above is called centralized HFL because it requires a trusted third party, a central server. On the other hand, decentralized HFL, which eliminates the need for a central server, has emerged in recent years \cite{beltran2023decentralized}. In decentralized HFL, clients directly communicates with each other, resulting in communication resource savings. 

% There are various possible methods of communication between clients \cite{beltran2023decentralized}. For example, the most common method for HFL of gradient boosting decision trees is for each client to add trees to the global model by sequence \cite{zhao2018inprivate,li2020practical,wang2021gradient}.

\subsection{Vertical Federated Learning}
VFL enables clients holding the different features of the same samples to collaboratively train a model that takes each client's various features as input. There are VFL studies that deal with various models including linear/logistic regression \cite{SLR,LR,LR2,LR3,LR4}, decision trees \cite{secureboost,tree,tree2,tree3,federboost}, neural networks \cite{fdml,NN,pyvertical,transnet}, and other non-linear models \cite{non-linear,fedv}.

\begin{figure}[tb]
\begin{center}
\includegraphics*[width=90mm]{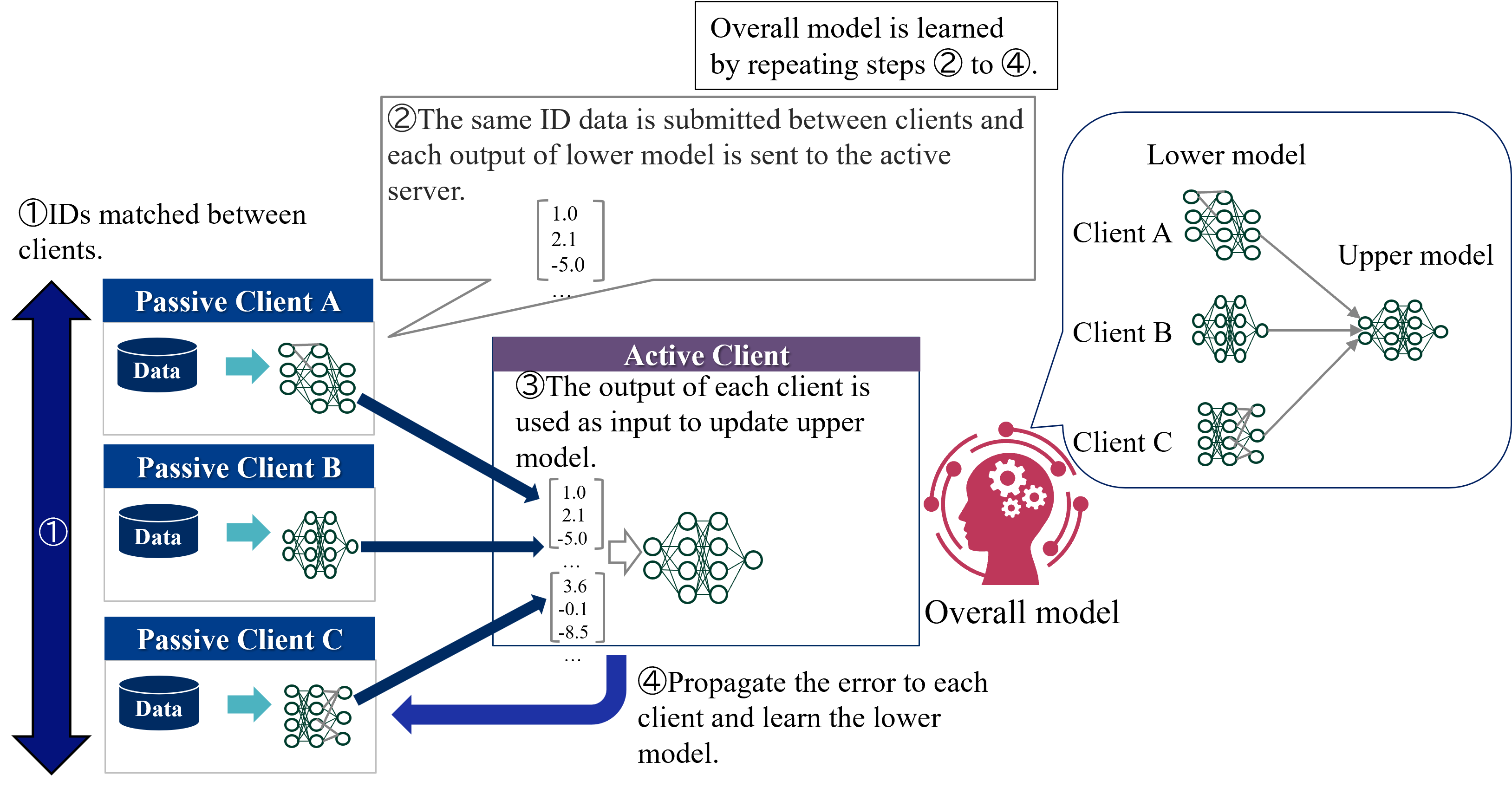}
\end{center}
\caption{Overview of the standard VFL learning protocol.}
\label{LM_VFL}
\end{figure}

Figure \ref{LM_VFL} shows an overview of the standard VFL learning protocol. In VFL, only one client holds labels, and it plays the role of a server. Therefore, two types of entities participate in learning of VFL:

\begin{enumerate}
    \renewcommand{\labelenumi}{\roman{enumi}}
    \item \textbf{Active client} - Features and labels owner. Active client coordinates the learning procedure. It calculates the loss and exchanges intermediate outputs with the passive clients.
    \item \textbf{Passive clients} - Features owners. Each passive client keeps both its features and model local but exchanges intermediate outputs with the active client.
\end{enumerate}
VFL consists of two phases: ID matching and learning phases. In the ID matching phases, all clients share the common sample IDs. In the learning phase, each client has a separate model with its own features as input, and the passive clients send the computed intermediate outputs to the active client. The active client calculates the loss based on the aggregated intermediate outputs and sends the gradients to all passive clients. Then, the passive clients update their own model parameters. This process is repeated until convergence. During inference time, all clients need to cooperate to predict the label of a sample.

\subsection{Federated Transfer Learning}
FTL assumes two clients that share only a small portion of samples or features. FTL aims to create a model that can predict labels on the client that does not possess labels (target client), by transferring the knowledge of the other client that does possess labels (source client) to the target client.

\begin{figure}[tb]
\begin{center}
\includegraphics*[width=90mm]{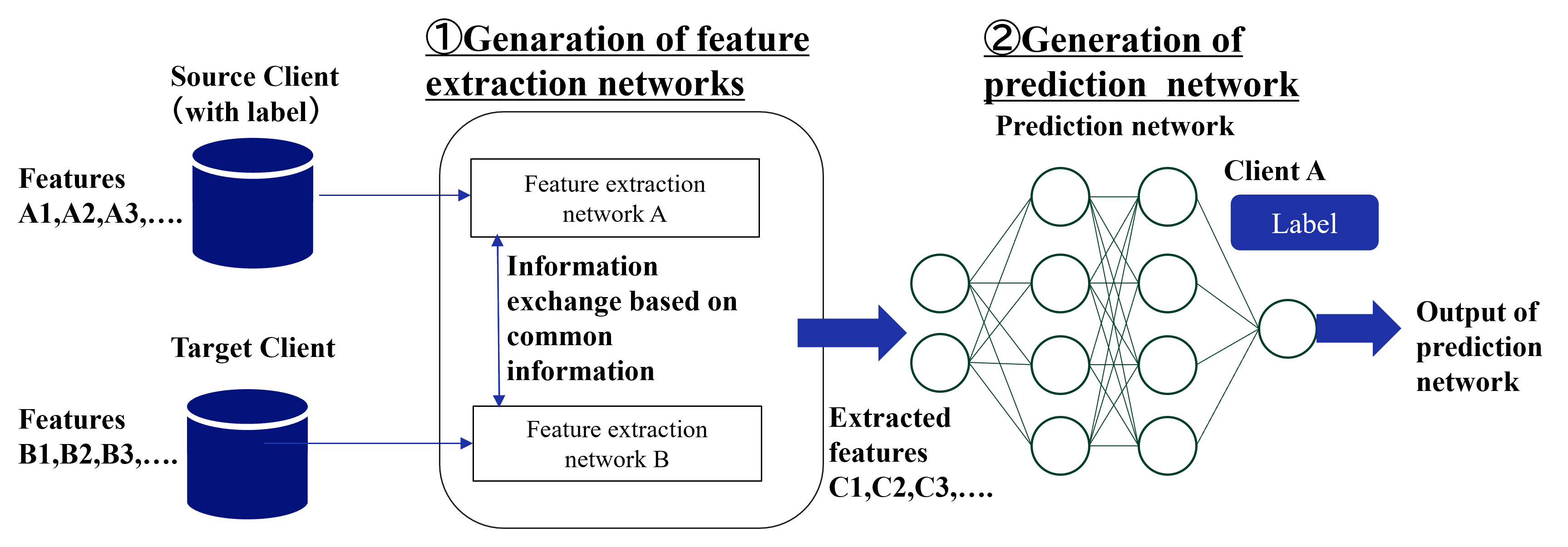}
\end{center}
\caption{Overall of the FTL learning protocol.}
\label{LM_FTL}
\end{figure}

Figure \ref{LM_FTL} shows an overall of the FTL learning protocol. As noted above, two types of entities participate in FTL:

\begin{enumerate}
    \renewcommand{\labelenumi}{\roman{enumi}}
    \item \textbf{Source client} - Features and labels owner. Source client exchanges intermediate outputs such as outputs and gradients with the target client and calculates the loss.
    \item \textbf{Target client} - Features owners. Target client exchanges intermediate outputs with the source client.
\end{enumerate}
In FTL, two clients exchange intermediate outputs to learn a common representation. The source client uses the labeled data to compute the loss and sends the gradient to the target client, which updates the target client's representation. This process is repeated until convergence. During inference, the target client predicts the label of a sample using its own model and features.

The detail of the learning protocol varies depending on the specific method. Although only a limited number of FTL methods have been proposed, we introduce three significant methods. FTL requires some supplementary information to bridge two clients, such as common IDs \cite{liu2020secure,sharma2019secure,zhang2020privacy,gao2022multiparty}, common features \cite{gao2019privacy,mori2022continual}, and labels of target client \cite{gao2019hhhfl,rakotomamonjy2023personalised}.

\subsubsection{ID-FTL}
Most FTL methods assume the existence of the common-ID samples between two clients. As with VFL, this type of FTL requires ID matching before the learning phase. Liu et al. \cite{liu2020secure} proposed the first FTL protocol, which learns feature transformation functions so that the different features of the common samples are mapped into the same features. Sharma et al. \cite{sharma2019secure} improved the communication overhead of the first FTL using multi-party computation and enhanced security by addressing malicious clients. Gao et al. \cite{gao2022multiparty} proposed a dual learning framework in which two clients impute each other's missing features by exchanging the outputs of their imputation models on the shared samples.

\subsubsection{Feature-FTL}
In real-world applications, sharing samples with the same IDs is difficult. Therefore, Gao et al. \cite{gao2019privacy} proposed a method to realize FTL by assuming common features instead of common samples. In that method, two clients mutually reconstruct the missing features by using exchanged feature mapping models. Then, using all features, the clients conduct HFL to obtain a label prediction model. In the original paper, the authors assume that all clients possess labels. However, this method applies to the target client that does not possess labels because the source client can only learn the label prediction model by itself. Mori et al. \cite{mori2022continual} proposed a method for neural networks in which each client incorporates its own unique features in addition to common features into HFL training. However, their method is based on HFL and cannot be applied to target clients who do not possess labels.

\subsubsection{Label-FTL}
This method assumes neither common IDs nor features but assumes that all clients possess labels, allowing a common representation to be learned across clients. Since it is based on HFL, the participating entities are the same as in HFL. Gao et al. \cite{gao2019hhhfl} learns a common representation by exchanging the intermediate outputs with the server and reducing maximum mean discrepancy loss. Rakotomamonjy et al. \cite{rakotomamonjy2023personalised} proposed a method to learn a common representation by using Wasserstein distance for intermediate outputs, which enables the clients only to exchange statistical information such as mean and variance with the server. 

\section{Threat Models and Privacy Attacks in Federated Learning}
% threat modelsで取り上げるべきものは？ これはどこに入れる？Table ? shows the privacy attacks.
This section shows the threat models and categorizes the privacy attacks. 

%% 説明の順番を以下のように変更

\subsection{Threat Models}
We define threat models based on four perspectives: FL type, attacker role, attack information, and attack style.

% 1. Attacker role
% server, client, third partyの3つを分かりやすくするために箇条書きで記載する。
% third partyは外部からアクセスしてくるpartyであると補足説明を加える。(third partyと書くだけで伝わらなさそう)
% 内部からの攻撃者であるserverとclientの場合の方が脅威である旨を述べ、本サーベイではthird partyの場合を省く旨を記載する。

\subsubsection{Attacker role}
We define three attacker roles: server, client, and third party. 
\begin{enumerate}
    \renewcommand{\labelenumi}{\roman{enumi}}
    \item \textbf{Server} - Denotes attacks initiated by the server. In VFL and FTL, an active client and a source client correspond to this role, respectively.
    \item \textbf{Client} - Denotes attacks initiated by a client. In VFL and FTL, this corresponds to a passive client and a target client, respectively.
    \item \textbf{Third Party} - Refers to attackers external to both server and client roles. 
\end{enumerate}

% 2. Information for attack
% 攻撃者がserverかclientなのかで使用できる情報が変わることを明記。
% serverなら各ラウンドで各クライアントからの勾配orモデルパラメータを手に入れることが出来、clientなら各ラウンドでグローバルモデルパラメータのみ使用可能。

\subsubsection{Information for attack}
% An attacker uses information to infer private data. The information is mostly gradient or model parameter. In terms of privacy, communicating gradients throughout the training process can reveal sensitive information and even cause deep leakage. Even a small portion of gradients can reveal a significant amount of sensitive information about the local data. From the shared local model parameter, the attacker can infer class representatives, dataset membership and properties, even the original training inputs and labels. The information that can be used by attackers depends on whether they are the server or the client.

An attacker can exploit shared information, such as gradients or model parameters, to infer private data. During training, exchanging gradients may expose sensitive information and lead to deep leakage, even when only a small subset is shared. Such gradients can reveal significant details about local data. From shared local model parameters, an attacker may infer class representatives, dataset membership, and characteristics or even reconstruct original training inputs and labels. The specific information accessible to an attacker depends on their role in the system:

\begin{enumerate}
    \renewcommand{\labelenumi}{\roman{enumi}}
    \item \textbf{Server} - Receives gradients or model parameters from each client in every round.
    \item \textbf{Client} - Has access only to model parameters in each round.
    \item \textbf{Third Party} -  May eavesdrop on communication and access gradients or model parameters.
\end{enumerate}

%% 3. Additional attacker knowledge
% もう少し色々なパターンを記載。
% 例1. MIAの場合は、member or non-memberの正解ラベルが振られたデータセットと、推論を行いたいデータの特徴量とラベル。
% 例2. attribute inferenceであれば、復元ターゲットの特徴量以外の特徴量。

%% 4. Attack style
% malicious, honest-but-curious以外に、honestも追加しておく？例えばID leakageの場合、honestの場合でもIDがばれてしまい問題になり得る。

\subsubsection{Attack style}
We define three attack styles: malicious, honest-but-curious, and honest.

\begin{enumerate}
    \renewcommand{\labelenumi}{\roman{enumi}}
    \item \textbf{Malicious} - Actively interferes with the training process to extract private information from the clients.
    \item \textbf{Honest-but-curious} - Passively follows FL protocols without disrupting training, but attempts to infer private information from received data.
    \item \textbf{Honest} - Passively follows FL protocols and does not attempt to infer private information from received data.
\end{enumerate}

%% 5. Federated learning type

\subsubsection{Federated learning type}
As explained in the previous section, we consider three types of FL: HFL, VFL, and FTL.
%% 6. Target of attack
% 後続の小節はthreat modelsに持ってくる。

\subsection{Privacy Attacks}
\begin{figure}[tb]
\begin{center}
\includegraphics*[width=74mm]{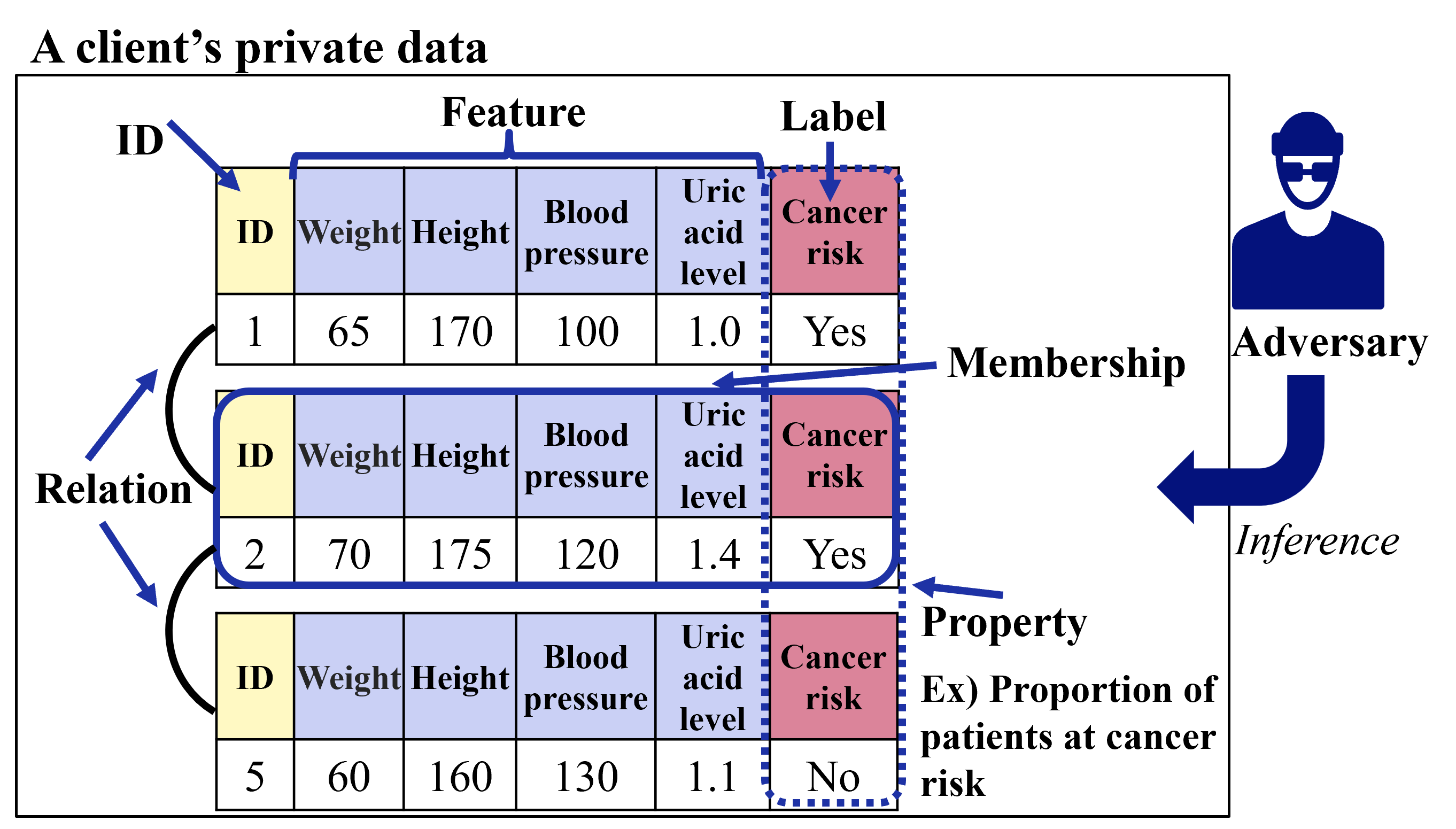}
\end{center}
\caption{Diagram of privacy attacks.}
\label{D_PA}
\end{figure}

We categorize privacy attacks into six types based on the nature of the data targeted: Feature Inference, Property Inference, Membership Inference, Label Inference, ID Leakage, and Relation Leaks. Figure~\ref{D_PA} illustrates a diagram of these categories, and Table~\ref{tab:threats} organizes them based on their corresponding threat models, along with relevant references addressing each threat. Although no existing studies specifically identify privacy threats in the context of FTL, we include "FTL" in the "FL type" column for those privacy threats potentially applicable to FTL. Each threat type is described below.

\subsubsection{Feature Inference}
Feature inference attacks aim to recover input features. If all features are targeted, the attack is also termed a reconstruction attack; if only a subset is targeted, it is also called attribute inference. The first such attack was introduced for HFL by \cite{NEURIPS2019_60a6c400}. More recently, various attacks have been developed for VFL. These attacks are especially effective in HFL settings using gradient boosting decision trees \cite{ito2024trojan}. HFL-based attacks apply to Feature-FTL and Label-FTL, which incorporate HFL in later stages, while VFL-based attacks are effective for ID-FTL, where common IDs are used.

\subsubsection{Property Inference}
Property inference attacks target properties of local datasets, such as class distributions \cite{wang2019eavesdropcompositionproportiontraining}, sensitive attributes \cite{zhang2021leakagedatasetpropertiesmultiparty}, or properties unrelated to the main task \cite{melis2018exploitingunintendedfeatureleakage}. These attacks are relevant in HFL, where client data distributions may differ. They typically assume that the adversary possesses auxiliary training data labeled with the target property \cite{melis2018exploitingunintendedfeatureleakage}.

% Property inference attack poses a threat in HFL, which has different property for each client.
% This attack estimates the property of data held by each client, such as class distributions \cite{wang2019eavesdropcompositionproportiontraining}, sensitive attribute distributions \cite{zhang2021leakagedatasetpropertiesmultiparty}, and some property independent of the properties that the global model aims to capture. In property inference attacks, it is assumed that the adversary has auxiliary training data that is correctly labeled with the target property \cite{melis2018exploitingunintendedfeatureleakage}.  

\subsubsection{Membership Inference}
Membership inference attacks attempt to determine whether specific data points were part of a client's training set. For instance, an attacker might infer whether a particular patient profile was used to train a disease classifier. These attacks are particularly effective in HFL, where each client has unique data samples and the server has access to model parameters, enabling powerful white-box attacks \cite{Nasr_2019}.

% In case of Membership Inference, the attack target is the membership of target data in some clients' training data. For example, an attacker may be able to infer whether a particular patient profile has been used to train a classifier that is associated with a particular disease. This attack is a threat in HFL cases where samples differ for each client and model inputs are common. In HFL, a server has access to model parameters, enabling more powerful white-box attacks to be proposed \cite{Nasr_2019}.

\subsubsection{Label Inference}
Label inference attacks aim to infer labels for samples when some clients lack label information. In VFL, passive clients do not possess labels but exchange intermediate computations, leading to potential label leakage. VFL architectures are inherently vulnerable to such attacks \cite{277244}. These attacks are also effective against ID-FTL, where clients share samples, but a target client does not possess labels.

% In case of Label Inference, the attack target is label. This attack is an attack in which clients that do not have labels estimate labels for common samples. In the VFL training process, passive clients do not have labels directly. Instead, they exchange intermediate computation results, posing a risk of label leakage. The current design of VFL has inherent vulnerabilities to label inference attacks \cite{277244}. These attack is also effective to ID-FTL, where both clients share some samples and a target client does not possess labels.

\subsubsection{ID Leakage}
In VFL and ID-FTL, which require alignment of shared samples across clients, there exists a risk that sample IDs held by one client may be exposed to others. Consequently, even honest clients may gain knowledge of the existence of entities that are either not part of the intersection \cite{hardy2017private} or that lie within the intersection \cite{sun2021verticalfederatedlearningrevealing} of shared samples. When there is asymmetry in sample IDs between clients, one client’s entire ID set is exposed to the other \cite{liu2020asymmetrical}, posing significant privacy risks. This inherent vulnerability severely limits the practical applicability of both VFL and ID-FTL.

% In VFL and ID-FTL, which require alignment of shared samples across clients, there exists a risk that the IDs of samples held by one client may become visible to other clients. Consequently, even honest clients may gain knowledge of the existence of entities that are either not part of the intersection \cite{hardy2017private} or that lie within the intersection \cite{sun2021verticalfederatedlearningrevealing} of shared samples. When there is asymmetry in the sample IDs held by each client, one client's entire set of IDs may be exposed to the other, posing a significant privacy risk \cite{liu2020asymmetrical}. This inherent vulnerability severly limits the practical applicability of VFL and ID-FTL.

\subsubsection{Relation Leaks}
In FL applied to relational data (e.g., graphs), relation leaks may occur, exposing connections between samples \cite{9899694}. These leaks exploit the intuition that related samples often have representations close in embedding space. A relationship can be inferred if the distance between two representations falls below a threshold.
% In FL applied to data with inter-sample relationships, such as graph data, there exists a potential risk of relation leakage, wherein the relationships between samples may be inadvertently revealed \cite{9899694}. Relation leaks are grounded in the intuition that the presence of a relation between two samples is strongly correlated with the distance between their corresponding representations. Specifically, if the distance between the representations of two samples falls below a certain threshold, it can be inferred that a relation exists between them.

\begin{table*}[tb]
    \centering
    \caption{Privacy threats and threat models}
    \begin{tabular}{|c|c|c|c|c|c|}
    \hline
    \textbf{Threat type} & \textbf{FL type} & \textbf{Reference} & \textbf{Attacker role} & \textbf{Information for attack} & \textbf{Attack style} 
     \\
    \hline
    \multirow{12}{*}{Feature Inference} & & \cite{10.1145/3510032},\cite{sun2021soteria},\cite{boenisch2023curiousabandonhonestyfederated},\cite{wei2021gradientleakageresilientfederatedlearning},\cite{lam2021gradientdisaggregationbreakingprivacy} & Server & Gradient & Malicious \\
    \cline{3-6}
    & & \cite{NEURIPS2019_60a6c400},\cite{zhao2020idlgimproveddeepleakage},\cite{10.5555/3495724.3497145},\cite{wei2020frameworkevaluatinggradientleakage} & \multirow{2}{*}{Server} & \multirow{2}{*}{Gradient} & \multirow{2}{*}{Honest-but-curious} \\
    & & \cite{vero2023tableaktabulardataleakage},\cite{10.5555/3540261.3540814},\cite{10025466},\cite{10231369} & & &  \\
    \cline{3-6}
    & HFL & \cite{wang2018inferringclassrepresentativesuserlevel},\cite{9109557} & Server & Model Parameter & Malicious \\
    \cline{3-6}
    & (Feature-FTL, Label-FTL) & \cite{hitaj2017deepmodelsganinformation},\cite{10.1145/3411501.3419423} & Client & Model Parameter  & Malicious \\
    \cline{3-6}
    & & \cite{9456909} & 3rd party & Model Parameter & Malicious \\
    \cline{3-6}
    & & \cite{10209197} & 3rd party & Gradient & Honest-but-curious \\
    \cline{3-6}
    & & \cite{li2020quantificationleakagefederatedlearning},\cite{bhowmick2019protectionreconstructionapplicationsprivate},\cite{So_Jiao_Avestimehr_2023} & Client & Model Parameter & Honest-but-curious \\
    \cline{2-6}
    & \multirow{4}{*}{VFL (ID-FTL)} & \cite{sun2021defendingreconstructionattackvertical},\cite{10386594},\cite{jiang2022comprehensive} & Server & Gradient & Malicious  \\
    \cline{3-6}
    & & \cite{10.5555/3540261.3540338} & Server & Gradient & Honest-but-curious  \\
    \cline{3-6}
    & & \cite{Luo_2021} & Server & Model Parameter & Honest-but-curious  \\
    \cline{3-6}
    & & \cite{weng2022practicalprivacyattacksvertical},\cite{Fu_2022},\cite{10122963} & Client & Model Parameter & Honest-but-curious \\
    \hline
    \multirow{4}{*}{Property Inference} & \multirow{4}{*}{HFL} & \cite{wang2019eavesdropcompositionproportiontraining},\cite{mo2021layerwisecharacterizationlatentinformation}& Server & Gradient & Malicious \\
    \cline{3-6}
    & & \cite{9210531} & Server & Model Parameter & Malicious  \\
    \cline{3-6}
    & & \cite{melis2018exploitingunintendedfeatureleakage},\cite{9204357},\cite{chase2021propertyinferencepoisoning} & Client & Model Parameter & Malicious  \\
    \cline{3-6}
    & & \cite{zhang2021leakagedatasetpropertiesmultiparty},\cite{9204357} & Client & Model Parameter & Honest-but-curious  \\
    \hline
    \multirow{4}{*}{Membership Inference} & \multirow{4}{*}{HFL} & \cite{Nasr_2019} & Server & Gradient & Malicious \\
    \cline{3-6}
    & & \cite{Nasr_2019} & Server & Gradient & Honest-but-curious \\
    \cline{3-6}
    & & \cite{truex2019demystifyingmembershipinferenceattacks},\cite{Nasr_2019},\cite{8927871},\cite{9209744} & Client & Model Parameter & Malicious \\
    \cline{3-6}
    & & \cite{Nasr_2019},\cite{9148790} & Client & Model Parameter & Honest-but-curious \\
    \hline
    \multirow{2}{*}{Label Inference} & \multirow{2}{*}{VFL (ID-FTL)} & \cite{277244},\cite{10024759} & Client & Model Parameter & Malicious \\
    \cline{3-6}
    & & \cite{10210670},\cite{sun2022labelleakageprotectionforward},\cite{takahashi2023eliminatinglabelleakagetreebased},\cite{li2022labelleakageprotectiontwoparty} & Client & Model Parameter & Honest-but-curious \\
    \hline
    \multirow{2}{*}{ID Leakage} & \multirow{2}{*}{VFL (ID-FTL)} & \cite{hardy2017private},\cite{liu2020asymmetrical} & Server, Client & Non-Aligned IDs & Honest  \\
    \cline{3-6}
    & & \cite{sun2021verticalfederatedlearningrevealing} & Server, Client & Aligned IDs & Honest \\
    \hline
    Relation Leaks & VFL (ID-FTL) & \cite{9899694} & Client & Model Parameter & Honest-but-curious \\
    \hline
    \end{tabular}
    \label{tab:threats}
\end{table*}

\section{General defense methods} %削除できる部分あり（森さん）
We summarize the defense methods against privacy attacks described in the previous section. These methods fall into two main categories: general methods, which are effective against most attacks, and specialized methods, which target specific attacks but are more efficient. While specialized methods are tailored to particular threats, they are often simpler and less computationally intensive than general ones, making them highly practical. This section focuses specifically on general defense methods.

Two widely used general approaches in privacy-preserving FL are:
(1) Communication channel defenses, which employ cryptographic techniques to protect model parameters or gradients and
(2) Differential privacy (DP), which prevents the leakage of data records from the model.

% We summarize the defence methods against privacy attacks described in the previous section. They are divided into two main categories: general defence methods, which are effective against almost all attacks, and specialized defence methods, which are effective only against certain attacks but are more convenient. These defence methods are specific to particular attacks, but are simpler and less computationally expensive than general defence methods. They are therefore highly practical. In particular, this section focuses on general defense methods.

% Two commonly used types of general defense methods in privacy-preserving FL studies are (1) \textit{communication channel defenses}, which use cryptographic techniques to protect model parameters or gradients, and (2) \textit{differential privacy (DP)}, which prevents the leakage of data records from the model.

\subsection{Communication Channel Defense}
These defenses prevent information leakage from shared raw intermediate outputs such as local model parameters, gradients, and outputs between the server and clients by encrypting them. However, they incur high computational costs. We review the two main techniques in this category: \textit{secure multi-party computation} (SMPC) and \textit{blockchain}. 

% These defenses allow us to prevent the information leakage from the shared raw intermediate outputs such as local model parameters, gradients, and outputs among the server and clients by encrypting the intermediate outputs. However, the computational cost is very high. We review the two main techniques in this categories: \textit{secure multi-party computation} (SMPC) and \textit{blockchain}. 

\subsubsection{Secure Multi-Party Computation (SMPC)}
SMPC \cite{smc} enables multiple parties to jointly compute a function over their private inputs without revealing them to each other. SMPC ensures that no party learns anything beyond the final output. SMPC is naturally suited for securely aggregating local model parameters and gradients in HFL, especially when the server is untrusted (malicious or honest-but-curious). In VFL and FTL, where more sensitive intermediate outputs must be shared, SMPC techniques are typically integrated into most proposed algorithms. However, while SMPC protects intermediate outputs, it does not safeguard the final aggregated output, which remains susceptible to inference attacks. Three primary methods to realize SMPC are: \textit{homomorphic encryption} (HE), \textit{garbled circuit} (GC), and \textit{secret sharing} (SS).

% SMPC \cite{smc}  allows multiple parties to compute a function over their private inputs  without revealing them to each other, which guarantees that all the parties cannot learn anything except the output. This can be naturally applicable for secure aggregation of local model parameters and gradients in HFL, which is effective against untrusted servers (malicious or honest-but-curious). In VFL and FTL, which require sharing the more private intermediate outputs, SMPC techniques are originally incorporated within most of the proposed algorithms. We note that although SMPC protects the privacy of intermediate outputs, it cannot protect privacy from the final aggregated results, which might be vulnerable to inference attacks. Three main methods are known to realize SMPC: \textit{homomorphic encryption} (HE), \textit{garbled circuit} (GC), and \textit{secret sharing} (SS).

\textbf{Homomorphic Encryption (HE).} HE is a type of encryption technique that enables computations such as addition and multiplication over encrypted inputs without decryption.
% This means that the computation is performed with the inputs in encrypted state, and only the final encrypted result needs to be decrypted, which protects the privacy and confidentiality of data. 
% We denote $Enc$ as a encryption function whihc maps the raw input $x$ to the encrypted input $Enc(x)$. The additive homomorphism and multiplicative homomorphism are described as follows, respectively:
% $$Enc(x_1) \oplus Enc(x_2) = Enc(x_1 + x_2)$$
% $$Enc(x_1) \otimes Enc(x_2) = Enc(x_1 \times x_2),$$
% where $\oplus$ and $\otimes$ represent addition and multiplication operations over encrypted data, respectively. 
For example, as cryptographic techniques, Paillier \cite{paillier} and modified El Gamal \cite{elgamal} possess additive homomorphism property, and El Gamal and  RSA \cite{rsa} possess multiplicative homomorphism property.
% 計算コストが非常に高いことを記載 
% HEを実際に使ってるFL論文を引用

% ↓修正
\textbf{Garbled Circuit (GC).} 
GC, first introduced by Yao \cite{garbled_circuits}, constructs Boolean circuits for secure two-party computation. Research on GC has primarily focused on improving performance and strengthening security. Security advancements address threats from both semi-honest and malicious adversaries. At the same time, efforts to optimize GC performance without compromising security continue to be an active research area.

% GC is first introduced by Yao \cite{garbled_circuits}, wherein a Boolean circuit is constructed to enable secure computation between two parties. The development of GC technology has been focused on two key aspects: performance and security enhancements. The security aspect of the GC is mainly reflected in its ability to provide protection against both semi-honest and malicious adversaries. Meanwhile, research on enhancing the performance of GC schemes while maintaining a similar level of security remains an active area of investigation.

% ↓修正
\textbf{Secret Sharing (SS).} 
SS is a cryptographic technique that distributes a secret among multiple participants so no individual holds the complete secret. The original secret can be reconstructed only when sufficient shares are combined. SS enables secure training of machine learning models without exposing raw data to any participant. The model is divided into shares and distributed, allowing reconstruction only when enough shares are collected.
% SS is a cryptographic technique used to distribute a secret among a group of participants in such a way that no single participant has access to the complete secret. 
% % Instead, the secret is divided into shares, and each participant is given a share of the secret. 
% Only when a sufficient number of shares are combined can the original secret be reconstructed. 
% % By dividing the data into shares, SS can ensure that no single participant has access to the complete data. 
% This makes it possible to train the machine learning model securely without disclosing the raw data to any participant. This can be done by dividing the model into shares and distributing them among the participants, so that the model can only be reconstructed when a sufficient number of shares are combined.

% ↓修正
\subsubsection{Blockchain} 
Blockchain is a distributed ledger that securely links an expanding list of records (blocks) using cryptographic hashes. It underpins most digital cryptocurrencies and enables peer-to-peer FL without relying on a potentially untrusted server, unlike SMPC and HE.

% Blockchain is a distributed ledger with growing lists of records (blocks) that are securely linked together via cryptographic hashes, supporting most digital cryptocurrencies. This allows peer to peer FL, unlike SMC and HE, without assuming the existence of a server which may be untrusted.

% ↓修正
\subsection{Differential Privacy (DP)}
FL protocols often incorporate additional DP techniques to prevent information leakage from both local updates and the global model. DP ensures that the inclusion or exclusion of any individual’s data has a limited impact on the output, thereby protecting sensitive information. This protection is achieved by adding calibrated random noise to the outputs, proportional to the maximum possible influence a single data point could have. Importantly, DP assumes that adversaries may possess arbitrary external knowledge. Several methods exist for integrating DP into FL to safeguard client data privacy.
% To guarantee the protection against information leakage not only from the local parameters or gradients but also from the global model, additional DP techniques are used in FL protocols. 
% DP allows retrieving information, rigorously bounding the harm caused to individuals whose sensitive data are stored in the database. Basically, it hides the presence of an individual in the database. To achieve this, DP adds random noise to the outputs. Such noise is
% calibrated to the magnitude of the largest contribution that can be made to the output by an individual. It is important to note that DP assumes
% that the adversary owns arbitrary external knowledge. There are various ways to apply DP to FL to protect the privacy of each client's data.

\subsubsection{Central Differential Privacy (CDP)}
CDP is the original DP definition. The following is the definition of DP.

\begin{dfn}
    For $\epsilon > 0$ and $0 \leq \delta < 1$, a randomized mechanism $\mathcal{M}$ is called $(\epsilon,\delta)$-differentially private if and only if for all $S \subseteq \rm{Range}(\mathcal{M})$ and for all adjacent datasets $D$ and $D'$, we have 
    $$\rm{Pr}[\mathcal{M}(D) \in S] \leq \exp(\epsilon) \cdot \rm{Pr}[\mathcal{M}(D') \in S] + \delta$$
\end{dfn}

In FL, CDP is realized by a trusted server that adds noise to aggregated outputs, such as the global model or gradients. This noise reduces the ability of individual clients to infer information about others. However, noise addition often leads to performance degradation. Moreover, CDP is ineffective against untrusted servers, as it relies on the server to apply the noise.
% In FL, CDP is realized by the trusted server adds the noise to the aggregated results such as global model or global gradients. This makes it difficult for each client to infer information about other clients from the global model. However, the addition of noise causes a noticeable performance degradation. In addition, CDP is not effective against untrustworthy servers because the servers serve to add noise.

\subsubsection{Local Differential Privacy (LDP)}
We can extend DP to the setting where the server is untrusted by each client, adding noise to its own intermediate outputs and sharing them with the server. This is realized with LDP. 
% The following is the definition of LDP.

% We can extend DP to the setting where the server is untrusted by each client adding noise to its own intermediate outputs and sharing them with the server. This is realized with LDP. The following is the definition of LDP.

% \begin{dfn}
%     For $\epsilon > 0$ and $0 \leq \delta < 1$, a randomized mechanism $\mathcal{M}$ is called $(\epsilon,\delta)$-local differentially private if and only if for all $o \in \rm{Range}(\mathcal{M})$ and for all inputs $x$ and $x'$, we have 
%     $$\rm{Pr}[\mathcal{M}(x) = o] \leq \exp(\epsilon) \cdot \rm{Pr}[\mathcal{M}(x') = o] + \delta$$
% \end{dfn}

LDP ensures that each client's privacy is protected from other clients and the server. However, LDP requires each client to add a sufficient amount of noise, so the total amount of noise added is enormous, resulting in a more significant performance degradation than DP.

\subsubsection{Distributed Differential Privacy (DDP)}
DDP bridges the gap between LDP and CDP by integrating cryptographic protocols to preserve individual privacy. DDP avoids relying on a trusted server and achieves better utility than LDP. Theoretically, DDP matches CDP in utility, as both add the same amount of noise.

DDP reflects that noise added to a statistic is distributed among clients. Implementations typically involve each participant applying a noise mechanism with reduced variance. These mechanisms require stable distributions to ensure the overall output is well-calibrated and cryptographic techniques to conceal intermediate outputs from clients.

% DDP bridges the gap between LDP and CDP while ensuring the privacy of each individual by combining with cryptographic protocols. Therefore, DDP avoids placing trust in any server and offers better utility than LDP. Theoretically, DDP offers the same utility as CDP, as the total amount of noise is the same.

% The notion of DDP reflects the fact that the required noise in the target statistic is sourced from multiple participants. Approaches to DDP that implement an overall additive noise mechanism by summing the same mechanism run at each participant (typically with less noise) necessitate mechanisms with stable distributions—to guarantee proper calibration of known end-to-end response distribution—and cryptography for hiding all but the final result from participants.

\subsubsection{Participant-level Differential Privacy (PDP)}
The DP technique described above prevents the leakage of individual client data but is ineffective against property inference attacks that target properties of each client's data. PDP mechanisms are required to address this limitation, which blur information at the client level. PDP achieves this by coordinating and adding noise to make individual clients indistinguishable.

% The DP described so far is a technique to prevent information leakage of individual data of each client. Therefore, it is not effective against attribute estimation attacks that favor the attributes possessed by the entire client. To address this, PDPs that blur information for each client are needed. PDP is achieved by coordinating and adding noise that makes the client indistinguishable.

\section{Specialized Defense methods}
These defense methods target specific attacks but are simpler and less computationally intensive than general methods, making them highly practical.
% These defence methods are specific to particular attacks, but are simpler and less computationally expensive than general defence methods. They are therefore highly practical. In the case of label inference and relation leaks, the general defenses mentioned in the previous section are effective.

\subsection{Feature Inference}
%何を載せる？ Large batch size
Using a large batch size is a potential defense against privacy attacks \cite{zhao2020idlgimproveddeepleakage,li2020quantificationleakagefederatedlearning}. Larger batches increase optimization complexity by introducing more variables. A linear programming method is employed to reconstruct full-batch data from gradients \cite{li2020quantificationleakagefederatedlearning}. However, as the number of constraints increases, solution time grows significantly, making such attacks computationally infeasible.

Another simple defense is dropout, a regularization technique that randomly deactivates neurons to reduce overfitting \cite{wang2019eavesdropcompositionproportiontraining,melis2018exploitingunintendedfeatureleakage}. Dropout weakens attacks by reducing the number of active gradients visible to adversaries. This stochastic feature removal further obfuscates sensitive data during training.

% In the case of feature inference, using a large batch size and is a possible defense against privacy attacks \cite{zhao2020idlgimproveddeepleakage,li2020quantificationleakagefederatedlearning}. Increasing the batch size increases the complexity of mitigating leakage during optimization because it introduces more variables to handle. Specifically, a linear programming method is used to derive the full batch data from the calculated gradients \cite{li2020quantificationleakagefederatedlearning}. It's worth noting that a larger number of constraints in the linear programming formulation results in a correspondingly longer solution time, making it an infeasible attack.

% Another simple mitigation technique is to employ Dropout, a popular regularization technique used to mitigate overfitting in neural networks \cite{wang2019eavesdropcompositionproportiontraining,melis2018exploitingunintendedfeatureleakage}.
% Dropout randomly deactivates activations between neurons. Random deactivations can weaken attacks because the adversary observes fewer gradients corresponding to active neurons. Dropout stochastically removes features at each collaborative training step.

\subsection{Property Inference}
%何を載せる？Dropout
For property inference, dropout is again a viable defense \cite{wang2019eavesdropcompositionproportiontraining,melis2018exploitingunintendedfeatureleakage}. Another approach is to share fewer gradient updates, whereby participants disclose only a fraction of their gradients per round \cite{melis2018exploitingunintendedfeatureleakage}. This reduces both communication costs and potential information leakage.

% In the case of property inference, one possible defense is to employ Dropout as shown in the defense for feature inference \cite{wang2019eavesdropcompositionproportiontraining,melis2018exploitingunintendedfeatureleakage}. 
% Another possible defense is sharing fewer gradient updates \cite{melis2018exploitingunintendedfeatureleakage}. Participants in federated learning could share only a fraction of their gradients during each update. This reduces communication overhead and, potentially, leakage, since the adversary observes fewer gradients.

\subsection{Membership Inference}
%何を載せる？ Hardening
Mitigation techniques effective in centralized machine learning can also be applied to local training in FL. For example, regularization, dropout, and distillation, commonly used to prevent overfitting, have also shown effectiveness against membership inference attacks. Moreover, integrating empirically validated defenses designed to improve robustness against white-box membership inference attacks \cite{kcd,280000} into client-side training can further reduce privacy risks.

\subsection{Label Inference}
Various defense strategies have been proposed to mitigate label inference threats in VFL, focusing on either obfuscating gradients or disrupting the correlation between shared representations and private labels. Techniques such as gradient perturbation \cite{li2022labelleakageprotectiontwoparty} and synthetic gradient generation \cite{10210670} have proven effective with minimal performance loss. 
Optimization-based approaches also reduce an adversary’s inference ability by minimizing the correlation between intermediate embeddings and private labels \cite{sun2022labelleakageprotectionforward, 10024759}.
Additionally, architecture-specific defenses have emerged, such as combining label differential privacy with post-processing and applying mutual information regularization to tree-based models \cite{takahashi2023eliminatinglabelleakagetreebased}.

\subsection{ID Leakage}
% dummy id, data augmentation
Private Set Intersection (PSI) is a standard cryptographic technique in VFL and ID-FTL to prevent leakage of non-overlapping ID information during ID alignment \cite{hardy2017private}. To support asymmetric ID alignment, an improved PSI variant has been introduced \cite{liu2020asymmetrical}. 
However, these techniques still reveal membership information within the intersection, compromising complete privacy. 
% However, even with these techniques, ID membership information within the intersection is still disclosed, thus failing to ensure complete privacy protection. 
To address this, dummy ID insertion has been proposed \cite{sun2021verticalfederatedlearningrevealing}, enabling clients to proceed without knowing which IDs are matched and enhancing privacy for shared IDs.

\section{Conclusion}

This survey shows that FL remains vulnerable to diverse and paradigm-specific privacy threats. Our taxonomy indicates that although HFL has been widely studied, particularly concerning feature and membership inference, defenses against property inference, such as participant-level differential privacy, remain underdeveloped. In contrast, VFL faces inherent risks, including label inference and ID leakage, which require more targeted countermeasures. Depending on the setting, FTL inherits vulnerabilities from HFL and VFL, but systematic studies of FTL-specific risks and defenses are lacking. Future research should prioritize mitigating VFL’s inherent vulnerabilities, conducting in-depth analyses of FTL-specific threats, and advancing unified threat models and benchmarking frameworks. Finally, developing lightweight, generalizable defenses that preserve utility while addressing multiple threats simultaneously will be essential for practical and secure FL deployment.

%ここにメッセージを追加 For example, authors could discuss the maturity of HFL and VFL solution and provide specific suggestions for future work or go deeper into how the unique challenges each scenario faces differ. The conclusion with a brief reflection on current research gaps (e.g., insufficient benchmarking, under-explored FL paradigms, real-world deployment challenges) and potential future directions (e.g., privacy-preserving FL with minimal utility loss, cross-paradigm defenses, or unified threat models). 

%サーベイから判明した知見を追記
% In this paper, we have described privacy threats and countermeasures for federated learning by categorizing threat models and privacy attacks because common and unique privacy threats among three typical types of federated learning (HFL, VFL, and FTL) have not been categorized and described comprehensively and specifically. This survey revealed that research on countermeasures against some privacy attacks, like feature inference, property inference, membership inference, and label inference, is progressing. On the other hand, it was found that research on countermeasures against privacy attacks like ID leakage and relation leaks has not been conducted so much. Further research on countermeasures for ID leakage and relation leaks is necessary in the future.

%We have summarized the general defense methods, which are effective against almost all attacks, and the specialized defense methods, which are effective against only certain attacks but are more convenient.

\section*{Acknowledgment}
This R\&D includes the results of "Research and development of optimized AI technology by secure data coordination (JPMI00316)" by the Ministry of Internal Affairs and Communications (MIC), Japan.

\bibliographystyle{IEEEtran}
\bibliography{IEEEexample}

\begin{thebibliography}{10}
\providecommand{\url}[1]{#1}
\csname url@samestyle\endcsname
\providecommand{\newblock}{\relax}
\providecommand{\bibinfo}[2]{#2}
\providecommand{\BIBentrySTDinterwordspacing}{\spaceskip=0pt\relax}
\providecommand{\BIBentryALTinterwordstretchfactor}{4}
\providecommand{\BIBentryALTinterwordspacing}{\spaceskip=\fontdimen2\font plus
\BIBentryALTinterwordstretchfactor\fontdimen3\font minus
  \fontdimen4\font\relax}
\providecommand{\BIBforeignlanguage}[2]{{%
\expandafter\ifx\csname l@#1\endcsname\relax
\typeout{** WARNING: IEEEtran.bst: No hyphenation pattern has been}%
\typeout{** loaded for the language `#1'. Using the pattern for}%
\typeout{** the default language instead.}%
\else
\language=\csname l@#1\endcsname
\fi
#2}}
\providecommand{\BIBdecl}{\relax}
\BIBdecl

\bibitem{lyu2020threats}
L.~Lyu, H.~Yu, X.~Ma, C.~Chen, L.~Sun, J.~Zhao, Q.~Yang, and P.~S. Yu,
  ``Privacy and robustness in federated learning: Attacks and defenses,''
  \emph{IEEE Transactions on Neural Networks and Learning Systems}, vol.~35,
  no.~7, pp. 8726--8746, 2024.

\bibitem{liu2024vertical}
Y.~Liu, Y.~Kang, T.~Zou, Y.~Pu, Y.~He, X.~Ye, Y.~Ouyang, Y.-Q. Zhang, and
  Q.~Yang, ``Vertical federated learning: Concepts, advances, and challenges,''
  \emph{IEEE Transactions on Knowledge and Data Engineering}, vol.~36, no.~7,
  pp. 3615--3634, 2024.

\bibitem{yang2019federated}
Q.~Yang, Y.~Liu, T.~Chen, and Y.~Tong, ``Federated machine learning: Concept
  and applications,'' \emph{ACM Trans. Intell. Syst. Technol.}, vol.~10, no.~2,
  2019.

\bibitem{mcmahan2017communication}
B.~McMahan, E.~Moore, D.~Ramage, S.~Hampson, and B.~A.~y. Arcas,
  ``{Communication-Efficient Learning of Deep Networks from Decentralized
  Data},'' in \emph{Proceedings of the 20th International Conference on
  Artificial Intelligence and Statistics}, ser. Proceedings of Machine Learning
  Research, vol.~54.\hskip 1em plus 0.5em minus 0.4em\relax PMLR, 2017, pp.
  1273--1282.

\bibitem{beltran2023decentralized}
E.~T. Martínez~Beltrán, M.~Q. Pérez, P.~M.~S. Sánchez, S.~L. Bernal,
  G.~Bovet, M.~G. Pérez, G.~M. Pérez, and A.~H. Celdrán, ``Decentralized
  federated learning: Fundamentals, state of the art, frameworks, trends, and
  challenges,'' \emph{IEEE Communications Surveys \& Tutorials}, vol.~25,
  no.~4, pp. 2983--3013, 2023.

\bibitem{SLR}
A.~Gasc{\'o}n, P.~Schoppmann, B.~Balle, M.~Raykova, J.~Doerner, S.~Zahur, and
  D.~Evans, ``Secure linear regression on vertically partitioned datasets,''
  \emph{IACR Cryptol. ePrint Arch.}, vol. 2016, p. 892, 2016.

\bibitem{LR}
S.~Hardy, W.~Henecka, H.~Ivey{-}Law, R.~Nock, G.~Patrini, G.~Smith, and
  B.~Thorne, ``Private federated learning on vertically partitioned data via
  entity resolution and additively homomorphic encryption,'' \emph{CoRR}, vol.
  abs/1711.10677, 2017.

\bibitem{LR2}
R.~Nock, S.~Hardy, W.~Henecka, H.~Ivey{-}Law, G.~Patrini, G.~Smith, and
  B.~Thorne, ``Entity resolution and federated learning get a federated
  resolution,'' \emph{CoRR}, vol. abs/1803.04035, 2018.

\bibitem{LR3}
S.~Yang, B.~Ren, X.~Zhou, and L.~Liu, ``Parallel distributed logistic
  regression for vertical federated learning without third-party coordinator,''
  \emph{CoRR}, vol. abs/1911.09824, 2019.

\bibitem{LR4}
Q.~Zhang, B.~Gu, C.~Deng, and H.~Huang, ``Secure bilevel asynchronous vertical
  federated learning with backward updating,'' \emph{Proceedings of the AAAI
  Conference on Artificial Intelligence}, vol.~35, no.~12, pp.
  10\,896--10\,904, May 2021.

\bibitem{secureboost}
K.~Cheng, T.~Fan, Y.~Jin, Y.~Liu, T.~Chen, D.~Papadopoulos, and Q.~Yang,
  ``Secureboost: A lossless federated learning framework,'' \emph{IEEE
  Intelligent Systems}, vol.~36, no.~6, pp. 87--98, 2021.

\bibitem{tree}
J.~Vaidya, C.~Clifton, M.~Kantarcioglu, and A.~S. Patterson,
  ``Privacy-preserving decision trees over vertically partitioned data,''
  \emph{ACM Trans. Knowl. Discov. Data}, vol.~2, no.~3, oct 2008.

\bibitem{tree2}
Y.~Wu, S.~Cai, X.~Xiao, G.~Chen, and B.~C. Ooi, ``Privacy preserving vertical
  federated learning for tree-based models,'' \emph{Proc. VLDB Endow.},
  vol.~13, no.~12, p. 2090–2103, jul 2020.

\bibitem{tree3}
Y.~Liu, Y.~Liu, Z.~Liu, Y.~Liang, C.~Meng, J.~Zhang, and Y.~Zheng, ``Federated
  forest,'' \emph{IEEE Transactions on Big Data}, pp. 1--1, 2020.

\bibitem{federboost}
Z.~Tian, R.~Zhang, X.~Hou, J.~Liu, and K.~Ren, ``Federboost: Private federated
  learning for {GBDT},'' \emph{CoRR}, vol. abs/2011.02796, 2020.

\bibitem{fdml}
Y.~Hu, D.~Niu, J.~Yang, and S.~Zhou, ``Fdml: A collaborative machine learning
  framework for distributed features,'' in \emph{Proceedings of the 25th ACM
  SIGKDD International Conference on Knowledge Discovery \& Data Mining}, ser.
  KDD '19.\hskip 1em plus 0.5em minus 0.4em\relax New York, NY, USA:
  Association for Computing Machinery, 2019, p. 2232–2240.

\bibitem{NN}
Y.~Liu, Y.~Kang, X.~Zhang, L.~Li, Y.~Cheng, T.~Chen, M.~Hong, and Q.~Yang, ``A
  communication efficient collaborative learning framework for distributed
  features,'' \emph{CoRR}, vol. abs/1912.11187, 2019.

\bibitem{pyvertical}
D.~Romanini, A.~J. Hall, P.~Papadopoulos, T.~Titcombe, A.~Ismail, T.~Cebere,
  R.~Sandmann, R.~Roehm, and M.~A. Hoeh, ``Pyvertical: A vertical federated
  learning framework for multi-headed splitnn,'' \emph{CoRR}, vol.
  abs/2104.00489, 2021.

\bibitem{transnet}
Q.~He, W.~Yang, B.~Chen, Y.~Geng, and L.~Huang, ``Transnet: Training
  privacy-preserving neural network over transformed layer,'' \emph{Proc. VLDB
  Endow.}, vol.~13, no.~12, p. 1849–1862, jul 2020.

\bibitem{non-linear}
B.~Gu, Z.~Dang, X.~Li, and H.~Huang, ``Federated doubly stochastic kernel
  learning for vertically partitioned data,'' in \emph{Proceedings of the 26th
  ACM SIGKDD International Conference on Knowledge Discovery \& Data
  Mining}.\hskip 1em plus 0.5em minus 0.4em\relax New York, NY, USA:
  Association for Computing Machinery, 2020, p. 2483–2493.

\bibitem{fedv}
R.~Xu, N.~Baracaldo, Y.~Zhou, A.~Anwar, J.~Joshi, and H.~Ludwig, ``Fedv:
  Privacy-preserving federated learning over vertically partitioned data,''
  \emph{CoRR}, vol. abs/2103.03918, 2021.

\bibitem{liu2020secure}
Y.~Liu, Y.~Kang, C.~Xing, T.~Chen, and Q.~Yang, ``A secure federated transfer
  learning framework,'' \emph{IEEE Intelligent Systems}, vol.~35, no.~4, pp.
  70--82, 2020.

\bibitem{sharma2019secure}
S.~Sharma, C.~Xing, Y.~Liu, and Y.~Kang, ``Secure and efficient federated
  transfer learning,'' in \emph{2019 IEEE International Conference on Big Data
  (Big Data)}, 2019, pp. 2569--2576.

\bibitem{zhang2020privacy}
B.~Zhang, C.~Chen, and L.~Wang, ``Privacy-preserving transfer learning via
  secure maximum mean discrepancy,'' \emph{arXiv preprint arXiv:2009.11680},
  2020.

\bibitem{gao2022multiparty}
Y.~Gao, M.~Gong, Y.~Xie, A.~K. Qin, K.~Pan, and Y.-S. Ong, ``Multiparty dual
  learning,'' \emph{IEEE Transactions on Cybernetics}, vol.~53, no.~5, pp.
  2955--2968, 2023.

\bibitem{gao2019privacy}
D.~Gao, Y.~Liu, A.~Huang, C.~Ju, H.~Yu, and Q.~Yang, ``Privacy-preserving
  heterogeneous federated transfer learning,'' in \emph{2019 IEEE International
  Conference on Big Data (Big Data)}, 2019, pp. 2552--2559.

\bibitem{mori2022continual}
J.~Mori, I.~Teranishi, and R.~Furukawa, ``Continual horizontal federated
  learning for heterogeneous data,'' in \emph{2022 International Joint
  Conference on Neural Networks (IJCNN)}, 2022, pp. 1--8.

\bibitem{gao2019hhhfl}
D.~Gao, C.~Ju, X.~Wei, Y.~Liu, T.~Chen, and Q.~Yang, ``Hhhfl: Hierarchical
  heterogeneous horizontal federated learning for electroencephalography,''
  \emph{arXiv preprint arXiv:1909.05784}, 2019.

\bibitem{rakotomamonjy2023personalised}
A.~Rakotomamonjy, M.~Vono, H.~J.~M. Ruiz, and L.~Ralaivola, ``Personalised
  federated learning on heterogeneous feature spaces,'' \emph{arXiv preprint
  arXiv:2301.11447}, 2023.

\bibitem{NEURIPS2019_60a6c400}
\BIBentryALTinterwordspacing
L.~Zhu, Z.~Liu, and S.~Han, ``Deep leakage from gradients,'' in \emph{Advances
  in Neural Information Processing Systems}, H.~Wallach, H.~Larochelle,
  A.~Beygelzimer, F.~d\textquotesingle Alch\'{e}-Buc, E.~Fox, and R.~Garnett,
  Eds., vol.~32.\hskip 1em plus 0.5em minus 0.4em\relax Curran Associates,
  Inc., 2019. [Online]. Available:
  \url{https://proceedings.neurips.cc/paper_files/paper/2019/file/60a6c4002cc7b29142def8871531281a-Paper.pdf}
\BIBentrySTDinterwordspacing

\bibitem{ito2024trojan}
K.~Ito, B.~Enkhtaivan, I.~Teranishi, and J.~Sakuma, ``Trojan attribute
  inference attack on gradient boosting decision trees,'' in \emph{2024 IEEE
  9th European Symposium on Security and Privacy (EuroS\&P)}, 2024, pp.
  542--559.

\bibitem{wang2019eavesdropcompositionproportiontraining}
\BIBentryALTinterwordspacing
L.~Wang, S.~Xu, X.~Wang, and Q.~Zhu, ``Eavesdrop the composition proportion of
  training labels in federated learning,'' 2019. [Online]. Available:
  \url{https://arxiv.org/abs/1910.06044}
\BIBentrySTDinterwordspacing

\bibitem{zhang2021leakagedatasetpropertiesmultiparty}
\BIBentryALTinterwordspacing
W.~Zhang, S.~Tople, and O.~Ohrimenko, ``Leakage of dataset properties in
  multi-party machine learning,'' 2021. [Online]. Available:
  \url{https://arxiv.org/abs/2006.07267}
\BIBentrySTDinterwordspacing

\bibitem{melis2018exploitingunintendedfeatureleakage}
\BIBentryALTinterwordspacing
L.~Melis, C.~Song, E.~D. Cristofaro, and V.~Shmatikov, ``Exploiting unintended
  feature leakage in collaborative learning,'' 2018. [Online]. Available:
  \url{https://arxiv.org/abs/1805.04049}
\BIBentrySTDinterwordspacing

\bibitem{Nasr_2019}
\BIBentryALTinterwordspacing
M.~Nasr, R.~Shokri, and A.~Houmansadr, ``Comprehensive privacy analysis of deep
  learning: Passive and active white-box inference attacks against centralized
  and federated learning,'' in \emph{2019 IEEE Symposium on Security and
  Privacy (SP)}.\hskip 1em plus 0.5em minus 0.4em\relax IEEE, May 2019.
  [Online]. Available: \url{http://dx.doi.org/10.1109/SP.2019.00065}
\BIBentrySTDinterwordspacing

\bibitem{277244}
\BIBentryALTinterwordspacing
C.~Fu, X.~Zhang, S.~Ji, J.~Chen, J.~Wu, S.~Guo, J.~Zhou, A.~X. Liu, and
  T.~Wang, ``Label inference attacks against vertical federated learning,'' in
  \emph{31st USENIX Security Symposium (USENIX Security 22)}.\hskip 1em plus
  0.5em minus 0.4em\relax Boston, MA: USENIX Association, Aug. 2022, pp.
  1397--1414. [Online]. Available:
  \url{https://www.usenix.org/conference/usenixsecurity22/presentation/fu-chong}
\BIBentrySTDinterwordspacing

\bibitem{hardy2017private}
S.~Hardy, W.~Henecka, H.~Ivey-Law, R.~Nock, G.~Patrini, G.~Smith, and
  B.~Thorne, ``Private federated learning on vertically partitioned data via
  entity resolution and additively homomorphic encryption,'' \emph{arXiv
  preprint arXiv:1711.10677}, 2017.

\bibitem{sun2021verticalfederatedlearningrevealing}
\BIBentryALTinterwordspacing
J.~Sun, X.~Yang, Y.~Yao, A.~Zhang, W.~Gao, J.~Xie, and C.~Wang, ``Vertical
  federated learning without revealing intersection membership,'' 2021.
  [Online]. Available: \url{https://arxiv.org/abs/2106.05508}
\BIBentrySTDinterwordspacing

\bibitem{liu2020asymmetrical}
Y.~Liu, X.~Zhang, and L.~Wang, ``Asymmetrical vertical federated learning,''
  \emph{arXiv preprint arXiv:2004.07427}, 2020.

\bibitem{9899694}
P.~Qiu, X.~Zhang, S.~Ji, T.~Du, Y.~Pu, J.~Zhou, and T.~Wang, ``Your labels are
  selling you out: Relation leaks in vertical federated learning,'' \emph{IEEE
  Transactions on Dependable and Secure Computing}, vol.~20, no.~5, pp.
  3653--3668, 2023.

\bibitem{10.1145/3510032}
\BIBentryALTinterwordspacing
H.~Ren, J.~Deng, and X.~Xie, ``Grnn: Generative regression neural network—a
  data leakage attack for federated learning,'' \emph{ACM Trans. Intell. Syst.
  Technol.}, vol.~13, no.~4, may 2022. [Online]. Available:
  \url{https://doi.org/10.1145/3510032}
\BIBentrySTDinterwordspacing

\bibitem{sun2021soteria}
J.~Sun, A.~Li, B.~Wang, H.~Yang, H.~Li, and Y.~Chen, ``Soteria: Provable
  defense against privacy leakage in federated learning from representation
  perspective,'' in \emph{Proceedings of the IEEE/CVF conference on computer
  vision and pattern recognition}, 2021, pp. 9311--9319.

\bibitem{boenisch2023curiousabandonhonestyfederated}
\BIBentryALTinterwordspacing
F.~Boenisch, A.~Dziedzic, R.~Schuster, A.~S. Shamsabadi, I.~Shumailov, and
  N.~Papernot, ``When the curious abandon honesty: Federated learning is not
  private,'' 2023. [Online]. Available: \url{https://arxiv.org/abs/2112.02918}
\BIBentrySTDinterwordspacing

\bibitem{wei2021gradientleakageresilientfederatedlearning}
\BIBentryALTinterwordspacing
W.~Wei, L.~Liu, Y.~Wu, G.~Su, and A.~Iyengar, ``Gradient-leakage resilient
  federated learning,'' 2021. [Online]. Available:
  \url{https://arxiv.org/abs/2107.01154}
\BIBentrySTDinterwordspacing

\bibitem{lam2021gradientdisaggregationbreakingprivacy}
\BIBentryALTinterwordspacing
M.~Lam, G.-Y. Wei, D.~Brooks, V.~J. Reddi, and M.~Mitzenmacher, ``Gradient
  disaggregation: Breaking privacy in federated learning by reconstructing the
  user participant matrix,'' 2021. [Online]. Available:
  \url{https://arxiv.org/abs/2106.06089}
\BIBentrySTDinterwordspacing

\bibitem{zhao2020idlgimproveddeepleakage}
\BIBentryALTinterwordspacing
B.~Zhao, K.~R. Mopuri, and H.~Bilen, ``idlg: Improved deep leakage from
  gradients,'' 2020. [Online]. Available:
  \url{https://arxiv.org/abs/2001.02610}
\BIBentrySTDinterwordspacing

\bibitem{10.5555/3495724.3497145}
J.~Geiping, H.~Bauermeister, H.~Dr\"{o}ge, and M.~Moeller, ``Inverting
  gradients - how easy is it to break privacy in federated learning?'' in
  \emph{Proceedings of the 34th International Conference on Neural Information
  Processing Systems}, ser. NIPS '20.\hskip 1em plus 0.5em minus 0.4em\relax
  Red Hook, NY, USA: Curran Associates Inc., 2020.

\bibitem{wei2020frameworkevaluatinggradientleakage}
\BIBentryALTinterwordspacing
W.~Wei, L.~Liu, M.~Loper, K.-H. Chow, M.~E. Gursoy, S.~Truex, and Y.~Wu, ``A
  framework for evaluating gradient leakage attacks in federated learning,''
  2020. [Online]. Available: \url{https://arxiv.org/abs/2004.10397}
\BIBentrySTDinterwordspacing

\bibitem{vero2023tableaktabulardataleakage}
\BIBentryALTinterwordspacing
M.~Vero, M.~Balunović, D.~I. Dimitrov, and M.~Vechev, ``Tableak: Tabular data
  leakage in federated learning,'' 2023. [Online]. Available:
  \url{https://arxiv.org/abs/2210.01785}
\BIBentrySTDinterwordspacing

\bibitem{10.5555/3540261.3540814}
Y.~Huang, S.~Gupta, Z.~Song, K.~Li, and S.~Arora, ``Evaluating gradient
  inversion attacks and defenses in federated learning,'' in \emph{Proceedings
  of the 35th International Conference on Neural Information Processing
  Systems}, ser. NIPS '21.\hskip 1em plus 0.5em minus 0.4em\relax Red Hook, NY,
  USA: Curran Associates Inc., 2024.

\bibitem{10025466}
A.~Hatamizadeh, H.~Yin, P.~Molchanov, A.~Myronenko, W.~Li, P.~Dogra, A.~Feng,
  M.~G. Flores, J.~Kautz, D.~Xu, and H.~R. Roth, ``Do gradient inversion
  attacks make federated learning unsafe?'' \emph{IEEE Transactions on Medical
  Imaging}, vol.~42, no.~7, pp. 2044--2056, 2023.

\bibitem{10231369}
H.~Liu, B.~Li, C.~Gao, P.~Xie, and C.~Zhao, ``Privacy-encoded federated
  learning against gradient-based data reconstruction attacks,'' \emph{IEEE
  Transactions on Information Forensics and Security}, vol.~18, pp. 5860--5875,
  2023.

\bibitem{wang2018inferringclassrepresentativesuserlevel}
\BIBentryALTinterwordspacing
Z.~Wang, M.~Song, Z.~Zhang, Y.~Song, Q.~Wang, and H.~Qi, ``Beyond inferring
  class representatives: User-level privacy leakage from federated learning,''
  2018. [Online]. Available: \url{https://arxiv.org/abs/1812.00535}
\BIBentrySTDinterwordspacing

\bibitem{9109557}
M.~Song, Z.~Wang, Z.~Zhang, Y.~Song, Q.~Wang, J.~Ren, and H.~Qi, ``Analyzing
  user-level privacy attack against federated learning,'' \emph{IEEE Journal on
  Selected Areas in Communications}, vol.~38, no.~10, pp. 2430--2444, 2020.

\bibitem{hitaj2017deepmodelsganinformation}
\BIBentryALTinterwordspacing
B.~Hitaj, G.~Ateniese, and F.~Perez-Cruz, ``Deep models under the gan:
  Information leakage from collaborative deep learning,'' 2017. [Online].
  Available: \url{https://arxiv.org/abs/1702.07464}
\BIBentrySTDinterwordspacing

\bibitem{10.1145/3411501.3419423}
\BIBentryALTinterwordspacing
X.~Xu, J.~Wu, M.~Yang, T.~Luo, X.~Duan, W.~Li, Y.~Wu, and B.~Wu, ``Information
  leakage by model weights on federated learning,'' in \emph{Proceedings of the
  2020 Workshop on Privacy-Preserving Machine Learning in Practice}, ser.
  PPMLP'20.\hskip 1em plus 0.5em minus 0.4em\relax New York, NY, USA:
  Association for Computing Machinery, 2020, p. 31–36. [Online]. Available:
  \url{https://doi.org/10.1145/3411501.3419423}
\BIBentrySTDinterwordspacing

\bibitem{9456909}
X.~Yuan, X.~Ma, L.~Zhang, Y.~Fang, and D.~Wu, ``Beyond class-level privacy
  leakage: Breaking record-level privacy in federated learning,'' \emph{IEEE
  Internet of Things Journal}, vol.~9, no.~4, pp. 2555--2565, 2022.

\bibitem{10209197}
H.~Liang, Y.~Li, C.~Zhang, X.~Liu, and L.~Zhu, ``Egia: An external gradient
  inversion attack in federated learning,'' \emph{IEEE Transactions on
  Information Forensics and Security}, vol.~18, pp. 4984--4995, 2023.

\bibitem{li2020quantificationleakagefederatedlearning}
\BIBentryALTinterwordspacing
Z.~Li, Z.~Huang, C.~Chen, and C.~Hong, ``Quantification of the leakage in
  federated learning,'' 2020. [Online]. Available:
  \url{https://arxiv.org/abs/1910.05467}
\BIBentrySTDinterwordspacing

\bibitem{bhowmick2019protectionreconstructionapplicationsprivate}
\BIBentryALTinterwordspacing
A.~Bhowmick, J.~Duchi, J.~Freudiger, G.~Kapoor, and R.~Rogers, ``Protection
  against reconstruction and its applications in private federated learning,''
  2019. [Online]. Available: \url{https://arxiv.org/abs/1812.00984}
\BIBentrySTDinterwordspacing

\bibitem{So_Jiao_Avestimehr_2023}
\BIBentryALTinterwordspacing
J.~So, R.~E.~Ali, B.~Güler, J.~Jiao, and A.~S. Avestimehr, ``Securing secure
  aggregation: Mitigating multi-round privacy leakage in federated learning,''
  \emph{Proceedings of the AAAI Conference on Artificial Intelligence},
  vol.~37, no.~8, pp. 9864--9873, Jun. 2023. [Online]. Available:
  \url{https://ojs.aaai.org/index.php/AAAI/article/view/26177}
\BIBentrySTDinterwordspacing

\bibitem{sun2021defendingreconstructionattackvertical}
\BIBentryALTinterwordspacing
J.~Sun, Y.~Yao, W.~Gao, J.~Xie, and C.~Wang, ``Defending against reconstruction
  attack in vertical federated learning,'' 2021. [Online]. Available:
  \url{https://arxiv.org/abs/2107.09898}
\BIBentrySTDinterwordspacing

\bibitem{10386594}
M.~N. Vu, T.~R. Jeter, R.~Alharbi, and M.~T. Thai, ``Active data reconstruction
  attacks in vertical federated learning,'' in \emph{2023 IEEE International
  Conference on Big Data (BigData)}, 2023, pp. 1374--1379.

\bibitem{jiang2022comprehensive}
X.~Jiang, X.~Zhou, and J.~Grossklags, ``Comprehensive analysis of privacy
  leakage in vertical federated learning during prediction,'' \emph{Proceedings
  on Privacy Enhancing Technologies}, 2022.

\bibitem{10.5555/3540261.3540338}
X.~Jin, P.-Y. Chen, C.-Y. Hsu, C.-M. Yu, and T.~Chen, ``Cafe: catastrophic data
  leakage in vertical federated learning,'' in \emph{Proceedings of the 35th
  International Conference on Neural Information Processing Systems}, ser. NIPS
  '21.\hskip 1em plus 0.5em minus 0.4em\relax Red Hook, NY, USA: Curran
  Associates Inc., 2024.

\bibitem{Luo_2021}
\BIBentryALTinterwordspacing
X.~Luo, Y.~Wu, X.~Xiao, and B.~C. Ooi, ``Feature inference attack on model
  predictions in vertical federated learning,'' in \emph{2021 IEEE 37th
  International Conference on Data Engineering (ICDE)}.\hskip 1em plus 0.5em
  minus 0.4em\relax IEEE, Apr. 2021. [Online]. Available:
  \url{http://dx.doi.org/10.1109/ICDE51399.2021.00023}
\BIBentrySTDinterwordspacing

\bibitem{weng2022practicalprivacyattacksvertical}
\BIBentryALTinterwordspacing
H.~Weng, J.~Zhang, X.~Ma, F.~Xue, T.~Wei, S.~Ji, and Z.~Zong, ``Practical
  privacy attacks on vertical federated learning,'' 2022. [Online]. Available:
  \url{https://arxiv.org/abs/2011.09290}
\BIBentrySTDinterwordspacing

\bibitem{Fu_2022}
\BIBentryALTinterwordspacing
F.~Fu, H.~Xue, Y.~Cheng, Y.~Tao, and B.~Cui, ``Blindfl: Vertical federated
  machine learning without peeking into your data,'' in \emph{Proceedings of
  the 2022 International Conference on Management of Data}, ser. SIGMOD/PODS
  ’22.\hskip 1em plus 0.5em minus 0.4em\relax ACM, Jun. 2022. [Online].
  Available: \url{http://dx.doi.org/10.1145/3514221.3526127}
\BIBentrySTDinterwordspacing

\bibitem{10122963}
R.~Yang, J.~Ma, J.~Zhang, S.~Kumari, S.~Kumar, and J.~J. P.~C. Rodrigues,
  ``Practical feature inference attack in vertical federated learning during
  prediction in artificial internet of things,'' \emph{IEEE Internet of Things
  Journal}, vol.~11, no.~1, pp. 5--16, 2024.

\bibitem{mo2021layerwisecharacterizationlatentinformation}
\BIBentryALTinterwordspacing
F.~Mo, A.~Borovykh, M.~Malekzadeh, H.~Haddadi, and S.~Demetriou, ``Layer-wise
  characterization of latent information leakage in federated learning,'' 2021.
  [Online]. Available: \url{https://arxiv.org/abs/2010.08762}
\BIBentrySTDinterwordspacing

\bibitem{9210531}
M.~Shen, H.~Wang, B.~Zhang, L.~Zhu, K.~Xu, Q.~Li, and X.~Du, ``Exploiting
  unintended property leakage in blockchain-assisted federated learning for
  intelligent edge computing,'' \emph{IEEE Internet of Things Journal}, vol.~8,
  no.~4, pp. 2265--2275, 2021.

\bibitem{9204357}
M.~Xu and X.~Li, ``Subject property inference attack in collaborative
  learning,'' in \emph{2020 12th International Conference on Intelligent
  Human-Machine Systems and Cybernetics (IHMSC)}, vol.~1, 2020, pp. 227--231.

\bibitem{chase2021propertyinferencepoisoning}
\BIBentryALTinterwordspacing
M.~Chase, E.~Ghosh, and S.~Mahloujifar, ``Property inference from poisoning,''
  2021. [Online]. Available: \url{https://arxiv.org/abs/2101.11073}
\BIBentrySTDinterwordspacing

\bibitem{truex2019demystifyingmembershipinferenceattacks}
\BIBentryALTinterwordspacing
S.~Truex, L.~Liu, M.~E. Gursoy, L.~Yu, and W.~Wei, ``Towards demystifying
  membership inference attacks,'' 2019. [Online]. Available:
  \url{https://arxiv.org/abs/1807.09173}
\BIBentrySTDinterwordspacing

\bibitem{8927871}
Y.~Mao, X.~Zhu, W.~Zheng, D.~Yuan, and J.~Ma, ``A novel user membership leakage
  attack in collaborative deep learning,'' in \emph{2019 11th International
  Conference on Wireless Communications and Signal Processing (WCSP)}, 2019,
  pp. 1--6.

\bibitem{9209744}
J.~Chen, J.~Zhang, Y.~Zhao, H.~Han, K.~Zhu, and B.~Chen, ``Beyond model-level
  membership privacy leakage: an adversarial approach in federated learning,''
  in \emph{2020 29th International Conference on Computer Communications and
  Networks (ICCCN)}, 2020, pp. 1--9.

\bibitem{9148790}
J.~Zhang, J.~Zhang, J.~Chen, and S.~Yu, ``Gan enhanced membership inference: A
  passive local attack in federated learning,'' in \emph{ICC 2020 - 2020 IEEE
  International Conference on Communications (ICC)}, 2020, pp. 1--6.

\bibitem{10024759}
W.~Xia, Y.~Li, L.~Zhang, Z.~Wu, and X.~Yuan, ``Cascade vertical federated
  learning towards straggler mitigation and label privacy over distributed
  labels,'' \emph{IEEE Transactions on Big Data}, pp. 1--14, 2023.

\bibitem{10210670}
K.~Fan, J.~Hong, W.~Li, X.~Zhao, H.~Li, and Y.~Yang, ``Flsg: A novel defense
  strategy against inference attacks in vertical federated learning,''
  \emph{IEEE Internet of Things Journal}, vol.~11, no.~2, pp. 1816--1826, 2024.

\bibitem{sun2022labelleakageprotectionforward}
\BIBentryALTinterwordspacing
J.~Sun, X.~Yang, Y.~Yao, and C.~Wang, ``Label leakage and protection from
  forward embedding in vertical federated learning,'' 2022. [Online].
  Available: \url{https://arxiv.org/abs/2203.01451}
\BIBentrySTDinterwordspacing

\bibitem{takahashi2023eliminatinglabelleakagetreebased}
\BIBentryALTinterwordspacing
H.~Takahashi, J.~Liu, and Y.~Liu, ``Eliminating label leakage in tree-based
  vertical federated learning,'' 2023. [Online]. Available:
  \url{https://arxiv.org/abs/2307.10318}
\BIBentrySTDinterwordspacing

\bibitem{li2022labelleakageprotectiontwoparty}
\BIBentryALTinterwordspacing
O.~Li, J.~Sun, X.~Yang, W.~Gao, H.~Zhang, J.~Xie, V.~Smith, and C.~Wang,
  ``Label leakage and protection in two-party split learning,'' 2022. [Online].
  Available: \url{https://arxiv.org/abs/2102.08504}
\BIBentrySTDinterwordspacing

\bibitem{smc}
O.~Goldreich, ``Secure multi-party computation,'' \emph{Manuscript. Preliminary
  version}, vol.~78, no. 110, pp. 1--108, 1998.

\bibitem{paillier}
P.~Paillier, ``Public-key cryptosystems based on composite degree residuosity
  classes,'' in \emph{Advances in Cryptology --- EUROCRYPT '99}, J.~Stern,
  Ed.\hskip 1em plus 0.5em minus 0.4em\relax Berlin, Heidelberg: Springer
  Berlin Heidelberg, 1999, pp. 223--238.

\bibitem{elgamal}
T.~Elgamal, ``A public key cryptosystem and a signature scheme based on
  discrete logarithms,'' \emph{IEEE Transactions on Information Theory},
  vol.~31, no.~4, pp. 469--472, 1985.

\bibitem{rsa}
\BIBentryALTinterwordspacing
R.~L. Rivest, A.~Shamir, and L.~Adleman, ``A method for obtaining digital
  signatures and public-key cryptosystems,'' \emph{Commun. ACM}, vol.~21,
  no.~2, p. 120–126, feb 1978. [Online]. Available:
  \url{https://doi.org/10.1145/359340.359342}
\BIBentrySTDinterwordspacing

\bibitem{garbled_circuits}
\BIBentryALTinterwordspacing
M.~Bellare, V.~T. Hoang, and P.~Rogaway, ``Foundations of garbled circuits,''
  in \emph{Proceedings of the 2012 ACM Conference on Computer and
  Communications Security}, ser. CCS '12.\hskip 1em plus 0.5em minus
  0.4em\relax New York, NY, USA: Association for Computing Machinery, 2012, p.
  784–796. [Online]. Available: \url{https://doi.org/10.1145/2382196.2382279}
\BIBentrySTDinterwordspacing

\bibitem{kcd}
R.~Chourasia, B.~Enkhtaivan, K.~Ito, J.~Mori, I.~T. ishi, and H.~Tsuchida,
  ``Knowledge cross-distillation for membership privacy,'' in \emph{Proc. Priv.
  Enhancing Technol.}, vol.~2, 2022, p. 362–377.

\bibitem{280000}
X.~Tang, S.~Mahloujifar, L.~Song, V.~Shejwalkar, M.~Nasr, A.~Houmansadr, and
  P.~Mittal, ``Mitigating membership inference attacks by {Self-Distillation}
  through a novel ensemble architecture,'' in \emph{31st USENIX Security
  Symposium (USENIX Security 22)}.\hskip 1em plus 0.5em minus 0.4em\relax
  Boston, MA: USENIX Association, Aug. 2022, pp. 1433--1450.

\end{thebibliography}

%\begin{IEEEbiography}[{\includegraphics[width=1in,height=1.25in,clip,keepaspectratio]{author1.jpg}}]{Masahiro Hayashitani} received the B.E., M.E., and Ph.D. degrees in information and computer science from Keio University, Tokyo, Japan, in 2005, 2007, and 2017, respectively. In 2007, he joined NEC Corporation, where he was involved in photonic networking research at the System Platforms Research Laboratories. He is currently at Secure System Platform Research Laboratories, NEC Corporation, where he is involved in standardization activities related to AI (ISO/IEC JTC 1/SC 42) and biometrics (ISO/IEC JTC 1/SC 37). 
%\end{IEEEbiography}

%\begin{IEEEbiography}[{\includegraphics[width=1in,height=1.25in,clip,keepaspectratio]{author2.jpg}}]{Junki Mori} received the M.E. degree in nuclear engineering from Kyoto University in 2020. In 2020, he joined NEC Corporation. He is currently at Secure System Platform Research Laboratories, NEC corporation, where he is involved in research of AI security.
%\end{IEEEbiography}

%\begin{IEEEbiography}[{\includegraphics[width=1in,height=1.25in,clip,keepaspectratio]{author3.jpg}}]{Isamu Teranishi} received D.E. degrees in electronic engineering in 2008 from Tokyo Institute of Technology. He joined NEC Corporation as a researcher of cryptography in 2002. He is currently at Secure System Platform Research Laboratories, NEC Corporation, where he is involved in research of AI security. 
%\end{IEEEbiography}

%\EOD

\end{document}